\documentclass[twoside]{article}
\usepackage{amsmath}
\usepackage{mathtools}
\usepackage{amsfonts}
\usepackage{bm} 
\usepackage{rotating}
\usepackage{tikz}
\usetikzlibrary{bayesnet}
\usepackage{tabularx,hhline}
\usepackage{array}
\usepackage{tabulary}
\newcolumntype{K}[1]{>{\centering\arraybackslash}p{#1}}
\usepackage{array}
\usepackage{multirow}
\usepackage{caption}
\usepackage{subcaption}
\usepackage{tabularx,booktabs} 
\usepackage{standalone}

\usepackage{times}  
\usepackage{listings}
\usepackage{fancyvrb}
\fvset{fontsize=\normalsize}

\usepackage{hyperref}

\usepackage[preprint]{aistats2026}
\newtheorem{definition}{Definition}
\newtheorem{theorem}{Theorem}
\usepackage[noend]{algorithmic}
\usepackage{algorithm}
\usepackage{mathrsfs}
\usepackage{url}
\usepackage{natbib}
\usepackage{bbm}
\usepackage{amssymb}
\usepackage{framed}
\usepackage{float}
\usepackage{mathrsfs}
\usepackage{amssymb}
\usepackage{bbm}
\usepackage{makecell} 
\usepackage{graphicx}
\usepackage{epsfig}
\usepackage{epstopdf}

\begin{document}

%

%

\twocolumn[

\aistatstitle{CauchyNet: Compact and Data-Efficient Learning using Holomorphic Activation Functions}



\aistatsauthor{Hong-Kun Zhang\textsuperscript{1,2}$^\dagger$ \And Xin Li\textsuperscript{3}$^*$ \And Sikun Yang\textsuperscript{4,3}$^\ddagger$  \And Zhihong Xia\textsuperscript{5}$^\S$}

\aistatsaddress{\textsuperscript{1}School of Sciences, Great Bay University, Dongguan 52300, China
\\\textsuperscript{2}Department of Mathematics and Statistics, University of
Massachusetts, Amherst\\\textsuperscript{3}Great Bay Institute for Advanced Study, Great Bay University, China
\\\textsuperscript{4}School of Computing and Information Technology, Great Bay University, Dongguan 52300, China
\\\textsuperscript{5}Department of Mathematics, Northwestern University, Evanston, IL, USA
\\$^\dagger$hzhang@umass.edu~~$^*$xinli2023@u.northwestern.edu~~$^\ddagger$sikunyang@gbu.edu.cn\\$^\S$Corresponding author: xia@math.northwestern.edu
}
]

\begin{abstract}
A novel neural network inspired by \emph{Cauchy's integral formula}, is proposed for function approximation tasks that include time series forecasting, missing data imputation, etc. 
Hence, the novel neural network is named CauchyNet. 
By embedding real-valued data into the complex plane, CauchyNet efficiently captures complex temporal dependencies, surpassing traditional real-valued models in both predictive performance and computational efficiency. 
Grounded in \emph{Cauchy's integral formula} and supported by the universal approximation theorem, CauchyNet offers strong theoretical guarantees for function approximation. 
The architecture incorporates complex-valued activation functions, enabling robust learning from \emph{incomplete} data while maintaining a compact parameter footprint and reducing computational overhead. 
Through extensive experiments in diverse domains, including transportation, energy consumption, and epidemiological data, CauchyNet consistently outperforms state-of-the-art models in predictive accuracy, often achieving a $50\%$ lower mean absolute error with fewer parameters. 
These findings highlight CauchyNet's potential as an effective and efficient tool for data-driven predictive modeling, particularly in resource-constrained and data-scarce environments. The code used to reproduce the results will be released upon the publication.
\end{abstract}

\section{Introduction}
Over the past decade, artificial intelligence (AI) has expanded dramatically, powering applications ranging from large language models (e.g., GPT-3, ChatGPT)~\citep{NEURIPS2020_1457c0d6,NIPS2017_3f5ee243}. However, the pursuit of ever-larger, more intricate models has raised significant concerns over financial and environmental costs~\citep{strubell-etal-2019-energy, patterson2021carbonemissionslargeneural, 10.1145/3442188.3445922}. These challenges are particularly pronounced for smaller research labs, edge devices, and low-power IoT platforms, which often cannot sustain the memory or energy demands of massive-scale architectures. 
A key issue arises in sequence-modeling tasks. While Transformers~\citep{NIPS2017_3f5ee243} excel at capturing long-range dependencies, their quadratic complexity with respect to sequence length limits their applicability in real-time or hardware-constrained scenarios. Similarly, Long Short-Term Memory (LSTM) networks~\citep{10.1162/neco.1997.9.8.1735} require significant parameter tuning and computational resources~\citep{Che2016RecurrentNN}, particularly when data are sparse.
Moreover, many real-world datasets are incomplete or partially observed, with curated corpora like M4~\citep{MAKRIDAKIS202054} being the exception~\citep{NAKAGAWA2008592, article_Adh}. 


Deep neural networks, including MLPs, CNNs, and RNNs, are known for their universal approximation
capability~\citep{10.5555/70405.70408}. However, these architectures often require large parameter budgets, particularly when dealing with functions that exhibit both data scarcity and high-magnitude oscillations~\citep{lecun2015deep,Goodfellow-et-al-2016}. 
To address these concerns,
this paper focuses on two key questions:
\begin{itemize}
    \item \emph{How can we design a neural network that remains lightweight in both computation and memory, while maintaining high accuracy in data-scarce scenarios?} 
    \item \emph{How can Cauchy's integral formula guide a universal approximation framework capable of handling near-singularities and partial data?}
\end{itemize} 

\begin{figure}
    \centering
     \includegraphics[width=0.45\textwidth]{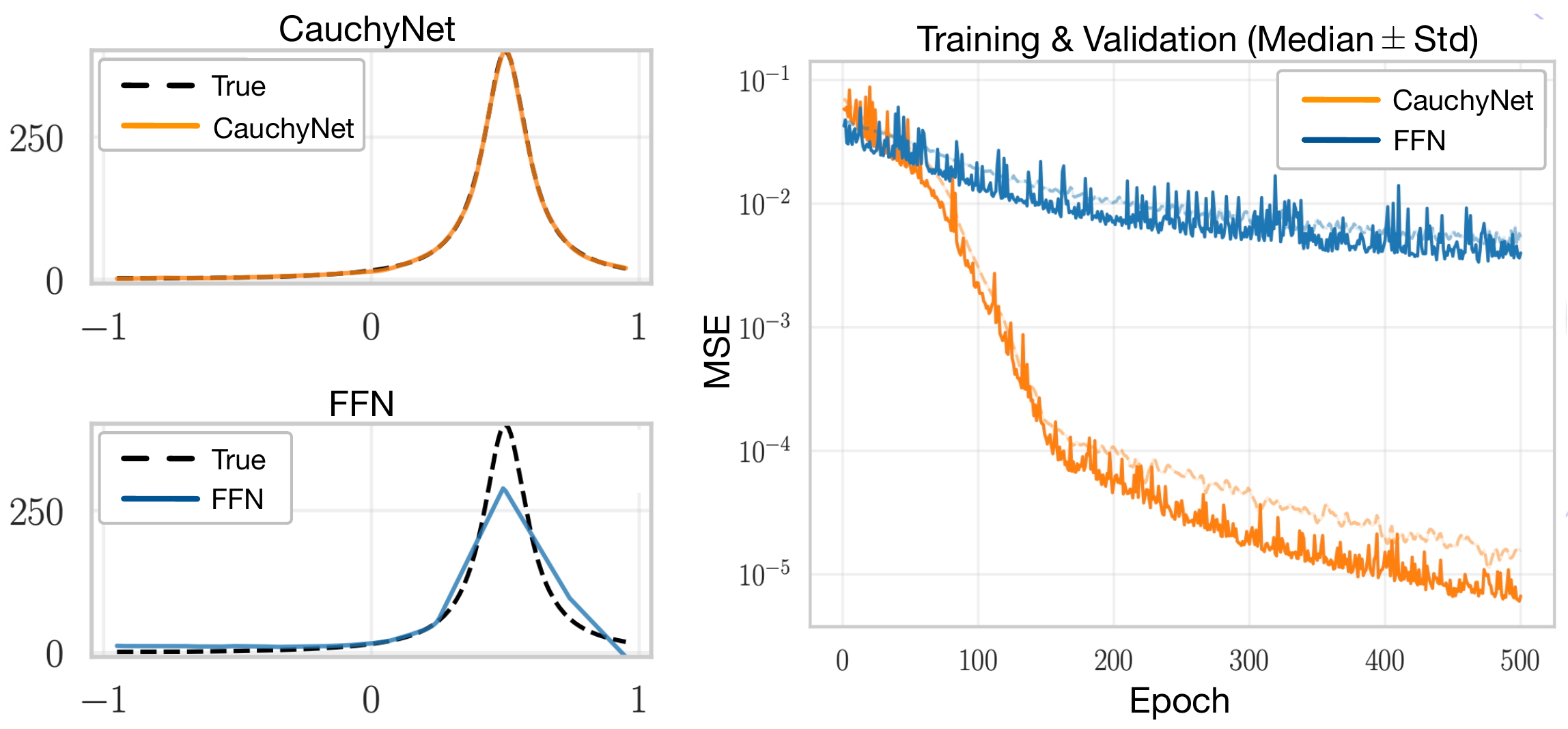}
    \caption{(Left) Comparison of the true function (dashed black line), CauchyNet (orange), and ReLU MLP (blue) in approximating a rational spike. The plot highlights CauchyNet's ability to closely track steep gradients, while the ReLU MLP significantly underestimates the peak near $x = 0.5$. (Right) Training and validation loss trajectories (log scale) over 500 epochs. Shaded regions indicate standard deviation across 10 independent runs, showing that CauchyNet converges faster and achieves lower final loss compared to the ReLU MLP.}
    \label{CauchyNetVsMLP}
\end{figure}

To the end, CauchyNet is introduced by leveraging complex analysis to achieve efficient function approximation with fewer parameters.
By incorporating an inversion-based activation, CauchyNet effectively represents abrupt changes and near-singularities with fewer neurons, compared to standard ReLU-based networks. The use of reciprocal transformations allows the network to capture both smooth oscillations and steep rational
spikes, common in real-world sensor data and physical simulations. This parameter-efficient design mitigates \emph{overfitting}, reducing sensitivity to high-magnitude outliers and noise. Moreover, the holomorphic framework of CauchyNet enables the use of Wirtinger derivatives, which satisfy $\partial f/\partial \bar{z} \equiv 0$ in ideal holomorphic settings. This property simplifies backpropagation and improves gradient stability, particularly in the presence of partial data or varying input scales~\citep{doi:10.1049/el:19921186}. 
To illustrate these advantages, consider the one-dimensional function:
\begin{align}
g(x)=\sin(3x)+{\frac{4}{(x-0.5)^{2}+0.01}},\quad x\in[-1,1],\notag
\end{align}
which exhibits a sharp, high-magnitude spike near $x = 0.5$. Conventional MLPs with ReLU activations struggle to approximate such peaks unless significantly over-parameterized, whereas our inversion-based activation naturally adapts to these near-singular behaviors. We conducted an empirical comparison between a ReLU-based MLP (FFN) and CauchyNet, each using a single hidden layer of 128 units. Both models were trained for 500 epochs using the Adam optimizer~\citep{2015-kingma} with an initial learning rate of 0.001, minimizing the mean squared error (MSE) over 200
training samples. No specialized regularization, aside from a mild imaginary penalty for CauchyNet, was applied. 
Fig.~\ref{CauchyNetVsMLP} presents the results. The left plots compare the predicted outputs of CauchyNet (orange) and FFN (blue) against the ground truth (dashed black). CauchyNet closely approximates the steep spike, whereas the FFN significantly underestimates its amplitude. The right plot shows the training and validation loss curves (log scale), where CauchyNet demonstrates faster convergence
and lower final loss, indicating its strong inductive bias for modeling rational-like near-singularities.
Beyond near-singularities, the holomorphic framework of CauchyNet extends naturally to handling incomplete data. The kernel-like inversion mechanism allows the network to interpolate effectively in missing regions while preserving structural consistency.

Overall, the proposed CauchyNet integrates a \emph{compact} parameterization with a \emph{holomorphic inductive bias}, achieving robust and efficient approximation capabilities even under challenging conditions such as data scarcity and abrupt changes. The theoretical motivation  is further validated by the empirical results ({Sec.~\ref{section_experiment} and the supplement}).

\section{Preliminary}
This section introduces the \emph{Cauchy's integral formula} underpins the proposed CauchyNet, and its role in kernel-based approximations. 
If $f$ is holomorphic on and within a closed contour $C \subset \mathbb{C}$, then for any $z$ inside $C$,
\begin{align}
f(z)=\frac{1}{2 \pi i} \oint_C \frac{f(\xi)}{\xi-z} d \xi.\notag    
\end{align}
This formula is driven by the inversion term $(\xi-z)^{-1}$, providing a basis for \emph{kernel-based expansions}. Extending to higher dimensions (i.e., $\mathbb{C}^N$) involves products of such inversions. More precisely, let $U$ be an open set in the complex space, such that
\begin{align}
U=\prod_{i=1}^N U_i \subset \mathbb{C}^N, \quad \text { with } \quad \bar{U}:=\prod_{i=1}^N \bar{U}_i \subset \mathbb{C}^N,\notag
\end{align}
where each $U_i$ is an open domain of $\mathbb{C}$, and $\bar{U}_i$ denotes its closure. The domain $U$ represents the Cartesian product of these $N$ sets, forming a multidimensional complex space. Let $M \subset U$. Suppose $f: M \rightarrow \mathbb{R}$ is extended to an analytic function $\bar{f}$ on $U$ and continuous on its closure $\bar{U}$. Then the high-dimensional Cauchy's integral formula is articulated for all $\boldsymbol{z}=\left(z_1, \cdots, z_N\right) \in \mathring{U}$:
\begin{align}
&\bar{f}(\boldsymbol{z})=
\frac{1}{(2 \pi i)^N}\times  \label{CauchyIntegralFormula}\\
&\int_{\zeta_1 \in \partial U_1} \cdots \int_{\zeta_N \in \partial U_N} \frac{\bar{f}(\zeta)}{\left(\zeta_1-z_1\right) \cdots\left(\zeta_N-z_N\right)} d \zeta_1 \ldots d \zeta_N,\notag
\end{align}
where $\zeta=\left(\zeta_1, \cdots, \zeta_N\right)$.

Indeed, we define in this paper a \emph{Cauchy Kernel} that can approximate functions over compact subsets of $\mathbb{R}^N$. A discrete analog emerges when sampling boundary points $\left\{\boldsymbol{\xi}_k\right\}$ and combining terms $\left(\xi_k-x\right)^{-1}$. This is central to the \emph{Cauchy Approximation Theorem}, see Sec.~\ref{UniApprox4CN} for details, ensuring that finite sums of such kernel evaluations suffice to approximate broad classes of real-valued targets. Our neural activation adopts this inversion directly,  instead of using piecewise-linear transformations, thus naturally addressing rational-like behaviors. \citet{Broomhead1988MultivariableFI}~also demonstrated that rational functions can effectively be employed for approximating complex functions.

\section{The proposed CauchyNet}
\begin{figure*}
    \centering
    \includegraphics[scale=0.08]{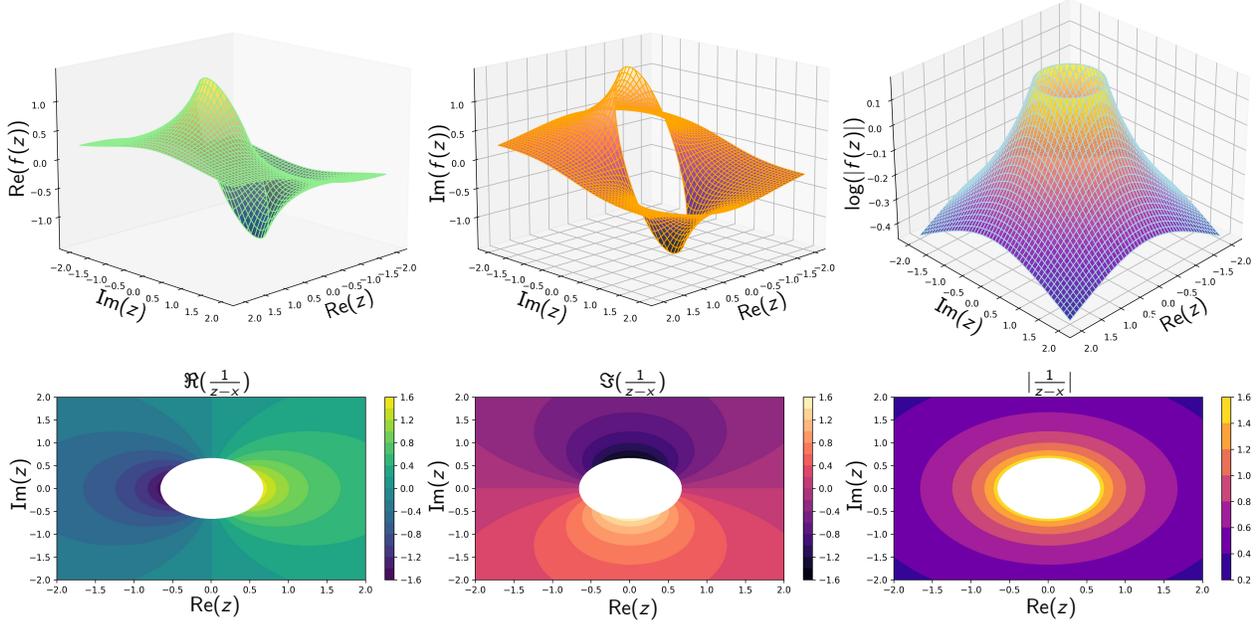}
    \caption{Visual representation of the Cauchy activation function $\mathscr{X}(\boldsymbol{z})$ for $z$ values outside a small
circular disc. (Top row) 3D surface plots showing the real part, imaginary part, and magnitude of $\mathscr{X}(\boldsymbol{z})$. (Bottom row) Corresponding 2D contour plots for each component. The white region in the
middle represents the excluded circular disc, demonstrating how the activation function transforms
inputs in the complex plane.}
    \label{VisCAF}
\end{figure*}
In this section, we introduce CauchyNet, a single-hidden-layer neural architecture that harnesses a Cauchy-inspired activation function. The primary objective is to achieve a lean parameter footprint, which is ideal for edge computing or data scarce environments, while maintaining robust capabilities to model oscillatory and near-singular phenomena.
\subsection{Cauchy Activation Function}
Inspired by \emph{Cauchy's integral formula}, we adopt an inversion-based activation function that excels in capturing near-singularities and oscillatory behaviors more efficiently than conventional activations like ReLU or Tanh.

\begin{definition}(Cauchy Activation Function).
Let $\mathbb{C}_*^N$ denote the set of all $N$-dimensional complex vectors with no zero components, defined as $\mathbb{C}_*^N=\left\{\boldsymbol{z}=\left(z_1, \cdots, z_N\right) \in \mathbb{C}^N: z_i \neq 0\right.$ for all $i= 1, \ldots, N\}$. 
\end{definition} 
The Cauchy activation function $\mathscr{X}$ is defined for any vector $z \in \mathbb{C}_*^N$ as:
\begin{align}
\mathscr{X}(\boldsymbol{z})=\prod_{i=1}^N z_i^{-1}.\notag
\end{align}
This activation function performs a component-wise multiplicative inversion, and aggregates the outcomes into a single complex number, effectively transforming complex vector inputs within the network. One notable property of this activation function is the simplicity of its derivative. For example, when $\boldsymbol{z} \in \mathbb{C}$ is a single complex number, the derivative can be expressed as:
\begin{align}
\frac{d}{d z} \mathscr{X}(z)=-\frac{1}{z^2}=-\mathscr{X}(z)^2.\notag
\end{align}
Fig.~\ref{VisCAF} illustrates the 3D plots of real, imaginary, and magnitude parts of the activation function $\mathscr{X}(\boldsymbol{z})$ for $z$ values taken outside of a small circular disc.

The multiplicative nature of the inversion-based activation, allows {CauchyNet} to efficiently model interactions between input dimensions, capturing complex multiplicative relationships inherent in oscillatory and near-singular phenomena. This stands in contrast to additive activations like ReLU, which may necessitate deeper architectures to achieve comparable representational power.

Furthermore, the inversion-based activation imparts a \emph{holomorphic inductive bias}, facilitating the approximation of complex-valued functions with desirable smoothness and boundedness properties. Moreover, this property allows the use of Wirtinger derivatives for gradient-based training, potentially stabilizing updates under partial or widely varying data~\citep{doi:10.1049/el:19921186}. The inversion inherently suppresses large values and amplifies small ones, aiding in the prevention of \emph{runaway gradients} or vanishing updates that can occasionally afflict ReLU-based networks.

\subsection{Overall Architecture and Parameterization.} 
CauchyNet is distinguished by its use of complex-valued parameters and an inversion-based activation function, which together enable efficient modeling of complex interactions within the data. The architecture comprises the following key components: the input embedding, the complex bias shifts, the new activation function, and the output combination.

\noindent\textbf{Forward Pass and Output Layer.} 
The forward pass of CauchyNet proceeds through the following stages:\\
1) \emph{Input Embedding into the Complex Plane}: An input vector $\boldsymbol{x} \in \mathbb{R}^m$ is embedded into the complex plane as $\boldsymbol{z} \in \mathbb{C}^m$ via
\begin{align}
\boldsymbol{z}=\boldsymbol{x}+i \mathbf{0},\notag
\end{align}
ensuring no imaginary components initially. This complex representation allows each hidden neuron to apply distinct bias shifts.\\
2) \emph{Complex Bias Shifts}: The embedded input is replicated across $h$ hidden neurons, forming $\mathbf{Z} \in \mathbb{C}^{h \times m}$. A learnable complex bias matrix $\mathbf{B} \in \mathbb{C}^{h \times m}$ is then added:
\begin{align}
\mathbf{H}=\mathbf{Z}+\mathbf{B},\notag
\end{align}
where each row $\mathbf{H}_k \in \mathbb{C}^m$ corresponds to the $k$-th hidden neuron.\\
3) \emph{Cauchy Activation}: Each hidden neuron applies the inversion-based activation function:
\begin{align}
h_k=\prod_{i=1}^m\left(H_{k, i}+\varepsilon\right)^{-1}, \quad k=1, \ldots, h.\notag
\end{align}
This yields the activation vector $\boldsymbol{h} \in \mathbb{C}^h$.\\
4) \emph{Output Combination}: The activated hidden units are aggregated using a complex coefficient vector $\mathbf{C} \in \mathbb{C}^{h \times 1}$ :
\begin{align}
o=\mathbf{C}^{\top} \boldsymbol{h},\label{activated_hidden_units}
\end{align}
resulting in a complex output $o=y+i e$, where $y=\Re(o)$ is the real-valued prediction, and $e=\Im(o)$ serves as an error term.

\begin{algorithm}[t]
\caption{Forward Pass of CauchyNet}\label{alg_FP}
\begin{algorithmic}[1]
\REQUIRE Input $\mathbf{x} \in \mathbb{R}^m$, Complex bias $\mathbf{B} \in \mathbb{C}^{h \times m}$, 
Complex coefficients $\mathbf{\Theta} \in \mathbb{C}^h$, offset $\varepsilon > 0$
\ENSURE Real output $y$, imaginary error $e$
\STATE $\mathbf{z} \gets \mathbf{x} + i\mathbf{0}$ \COMMENT{Embed into $\mathbb{C}^m$}
\STATE $\mathbf{Z} \gets$ Replicate $\mathbf{z}$ across $h$ rows
\STATE $\mathbf{H} \gets \mathbf{Z} + \mathbf{B}$ \COMMENT{Apply complex bias shifts}
\FOR{$k = 1$ \textbf{to} $h$}
    \STATE $h_k \gets \prod_{i=1}^m (H_{k,i} + \varepsilon)^{-1}$ \COMMENT{Cauchy Activation}
\ENDFOR
\STATE $\mathbf{h} \gets (h_1, \ldots, h_h)^\top$
\STATE $o \gets \mathbf{C}^\top h$
\STATE $y \gets \Re(o)$
\STATE $e \gets \Im(o)$
\RETURN $(y, e)$
\end{algorithmic}
\end{algorithm}

\noindent\textbf{Backward Pass and Gradient Computation.} 
While Algorithm~\ref{alg_FP} delineates the forward pass, the backward pass leverages Wirtinger derivatives
to compute gradients with respect to the complex parameters $\mathbf{B}$ and $\mathbf{C}$. This approach treats the
real and imaginary parts of complex variables separately, ensuring stable and efficient gradient
updates. However, modern deep learning frameworks like PyTorch inherently handle complex-valued
computations, simplifying the implementation of gradient-based optimization.

\begin{algorithm}[ht]
\caption{Backward Pass of CauchyNet}
\begin{algorithmic}[1]
\REQUIRE {Gradients of loss with respect to $y$ and $e$, Complex bias $\mathbf{B}$, 
      Complex coefficients $\mathbf{C}$, Hidden activations $\mathbf{h}$}
\ENSURE {Gradients with respect to $\mathbf{B}$ and $\mathbf{C}$}
\STATE Compute $\frac{\partial \mathcal{L}}{\partial o} 
= \frac{\partial \mathcal{L}}{\partial y} 
+ i \frac{\partial \mathcal{L}}{\partial e}$ \;
\STATE Compute gradients w.r.t. $\mathbf{C}$: 
$\frac{\partial \mathcal{L}}{\partial \mathbf{C}} 
= h \cdot \frac{\partial \mathcal{L}}{\partial o}$ \;
\FOR{$k = 1$ \textbf{to} $h$}{
    \STATE Compute $\frac{\partial \mathcal{L}}{\partial h_k} 
    = \mathbf{C}_k \cdot \frac{\partial \mathcal{L}}{\partial o}$ \;
    \STATE Compute $\frac{\partial \mathcal{L}}{\partial B_{k,i}} 
    = \frac{\partial \mathcal{L}}{\partial h_k} \cdot (-1) 
      \cdot \prod_{\substack{j=1 \\ j \neq i}}^m (H_{k,j} + \varepsilon)^{-1} 
      \cdot (H_{k,i} + \varepsilon)^{-2}, 
      \quad \forall i = 1, \ldots, m$ \;
}
\ENDFOR
\RETURN {Gradients w.r.t. $\mathbf{B}$ and $\mathbf{C}$}
\end{algorithmic}
\end{algorithm}

\noindent\textbf{Training Objective and Complexity.} 
Let $y_{\text {true }}$ denote the target output. The model produces a complex output\\ $o=y+i e$, where $y$ is the real-valued prediction, and $e=\Im(o)$ is the imaginary error term, see Eq.~\ref{activated_hidden_units}. To enforce accurate real-valued predictions, we define the loss function as:
\begin{align}
\mathcal{L}=\left(y-y_{\text {true }}\right)^2+\lambda|e|^2,
\end{align}
where $\lambda>0$ is a hyperparameter (typically set to $\lambda=0.1$ ) that penalizes the magnitude of the imaginary component, encouraging $e \approx 0$ during training.

Using Wirtinger derivatives, the loss gradients with respect to the complex parameters $\mathbf{B}$ and $\mathbf{C}$ are computed efficiently, allowing for seamless integration with gradient-based optimization algorithms. Modern frameworks such as PyTorch and TensorFlow handle complex-valued backpropagation intrinsically, streamlining the training process.

Each forward and backward pass of CauchyNet, incurs a computational complexity of $\mathcal{O}(h m)$, where $h$ is the number of hidden units, and $m$ is the input dimensionality, as detailed in Eq.~\ref{realparameters}.
\begin{align}
\underbrace{2(h \times m)}_{\text {Complex Biases B}}+\underbrace{2 h}_{\text {Complex Coefficients} \Theta}
=\underbrace{2h(m+1)}_{\text {(Real Parameters)}}.\label{realparameters}
\end{align}
This is considerably more efficient compared to recurrent or attention-based architectures, particularly when $h$ or $m$ scales moderately, or when real-time inference is required on resource-constrained hardware. This parameter efficiency is significantly lower than that of traditional architectures like Long Short-Term Memory (LSTM) networks or Transformer models, rendering CauchyNet particularly suitable for deployment in hardware-constrained or data-limited scenarios.

\noindent\textbf{Parameter Initialization Strategies.} We provide two strategies for parameter initialization. The first one is rather general.
The complex biases $\mathbf{B} \in \mathbb{C}^{h \times m}$ and complex coefficients $\mathbf{C} \in \mathbb{C}^{h \times 1}$ are initialized using a variant of the Xavier (Glorot) initialization scheme adapted for complex-valued parameters. Specifically, both the real and imaginary components are sampled from a normal distribution with zero mean and variance $\frac{2}{m+h}$:
\begin{align}
&\mathbf{B}_{k, i} \sim \mathcal{N}\left(0, \frac{2}{m+h}\right)+i \mathcal{N}\left(0, \frac{2}{m+h}\right), \\
&\mathbf{C}_k \sim \mathcal{N}\left(0, \frac{2}{m+h}\right)+i \mathcal{N}\left(0, \frac{2}{m+h}\right), \quad \forall k, i
\end{align}

This initialization strategy ensures that the variance of activations remains consistent across layers, promoting stable and efficient training dynamics. Inspired by \emph{Cauchy's integral formula} (Eq.~\ref{CauchyIntegralFormula}), we can also use the elliptical initialization strategy, see details in {Experiment 5 in the Supplement}.

\section{Related Work}
Our investigation addresses two main challenges outlined in \textcolor{black}{Sec. 1}, including the design of resource-efficient neural models for data-scarce environments, and the use of holomorphic techniques
for near-singular approximation and missing-data imputation.
From a theoretical perspective, rational expansions have been shown to provide strong approximations near singularities~\citep{Broomhead1988MultivariableFI,Park1991UniversalAU}. Although wavelets, radials and Sinusoidal~\citep{NEURIPS2020_53c04118, vonderfecht2024predictingencodingerrorsirens} predicting capture localized
features well, they are less capable of modeling rational-like spikes or high-magnitude gradients. In contrast, inversion-based expansions derived from the Cauchy kernel naturally handle abrupt peaks and facilitate interpolation in regions with missing data. Recent work has also explored complex-valued networks for oscillatory signals~\citep{doi:10.1049/el:19921186,NEURIPS2020_53c04118}. For instance, \citet{NEURIPS2020_53c04118} proposes implicit neural
representations that use periodic activation functions to capture high-frequency signals effectively, while Vonderfecht and Liu~\citep{vonderfecht2024predictingencodingerrorsirens} analyze the encoding error in such models. However, these approaches typically rely on multi-layer architectures with high parameter counts. By contrast, our single-layer, holomorphic-inspired CauchyNet is specifically designed to address near-singular phenomena with far fewer parameters, a distinct advantage when data are sparse or incomplete. Furthermore, a comprehensive survey~\citep{hammad2024comprehensivesurveycomplexvaluedneural} on complex-valued neural networks provides detailed insights into backpropagation techniques and the design of various activation functions in complex-valued settings. While the survey summarizes a wide range of existing activation strategies, our work distinguishes itself by introducing an inversion-based activation derived directly from \emph{Cauchy's integral formula}. This novel activation function yields a \emph{holomorphic inductive bias} within a \emph{compact}, single-layer architecture, enabling efficient function approximation even under challenging conditions such as near-singularities and data scarcity.
Meanwhile, a growing body of work have focused on enhancing forecasting performance and resource efficiency in edge settings~\citep{NBEATS,bandara2021mstlseasonaltrenddecompositionalgorithm}. These studies further motivate the development of a lightweight-yet-robust model like CauchyNet. 
In summary, while traditional models such as LSTMs, Transformer, and deep MLPs exhibit considerable power, they are often unsuitable for data-scarce or resource-constrained environments. 
This review motivates the need for an architecture that is both efficient and capable of handling near-singularities—an approach that underpins the design of CauchyNet.

\section{Theoretical Analysis}
\begin{figure}
    \centering
    \includegraphics[scale=0.65]{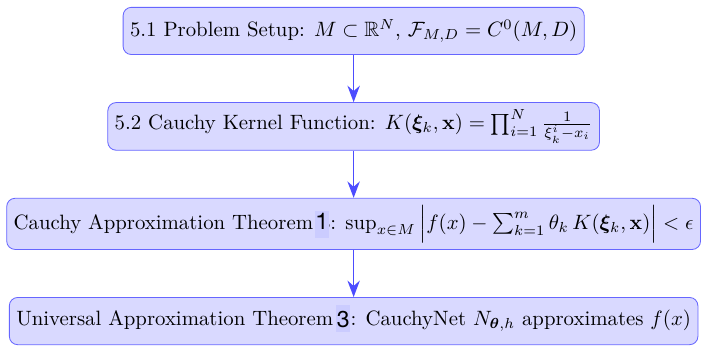}
    \caption{Roadmap of the theoretical analysis for CauchyNet}
    \label{flowchart}
 \end{figure}

This section formalizes why CauchyNet can approximate a broad class of functions defined on compact subsets of $\mathbb{R}^N$. The approach is based on complex analysis, specifically using a Cauchy-type kernel that leverages inversion-like terms from \emph{Cauchy's integral formula}. We first describe the general problem of continuous function approximation, and then show how CauchyNet inherits a universal approximation property through its holomorphic design. Fig.~\ref{flowchart} presents the main steps in the theoretical derivation.
\subsection{Problem Setup and Function Spaces}
Let
\[
	M=\prod_{i=1}^{N}M_{i}\subset\mathbb{R}^{N}
\]
be a compact domain (with each $M_{i}\subset\mathbb{R})$, and define
\[
	{\mathcal{F}}_{M,D}:=C^{0}(M,D)=\left\{f:M\rightarrow D\mid f{\mathrm{~is~continuous}}\right\},
\]
where $D=\mathbb{R}$ or $D=\mathbb{C}.$ Given a set of observed training data
\[
	{\cal T}:=\{({\bf x}_{i},{\bf y}_{i})\}_{i=1}^{n},
\]
with $\mathbf{y}_{i}=f(\mathbf{x}_{i})$ for an unknown dynamical system $f\in{\mathcal{F}}_{M,\mathbb{R}},$ our goal is to find a complex-valued function $g\in{\mathcal{F}}_{M,\mathbb{C}}$ that approximates  
$f$ on these samples. In practice, we minimize an empirical loss such as
\[
	L_{f,T}(g):=\frac{1}{n}\sum_{({\bf x}_{i},{\bf y}_{i})\in{\cal T}}\Big(|\Re(g({\bf x}_{i}))-{\bf y}_{i}|^{2}+\lambda\Big|\Im(g({\bf x}_{i}))\Big|^{2}\Big),
\]
where $\Re(g(\mathbf{x}))$ denotes the real part of $g(\mathbf{x})$, and
$\Im(g(\mathbf{x}))$ denotes the imaginary part. This loss encourages
$g$ to produce nearly real-valued predictions. 

\subsection{Cauchy Kernel
Function}
To bridge real-valued approximation with holomorphic kernels, we embed
${M}$ into an open domain
\[
	U=\prod_{i=1}^{N}U_{i}\subset\mathbb{C}^{N},
\]
with each $\ U_{i}$ chosen symmetric about the real axis and containing $M_{i}$. 
Let $\partial{\bar{U}}$ denote the boundary of $U$ (or its closure).

\begin{definition}{\rm(Cauchy Kernel Function).}  
Select $m$ distinct points
$\boldsymbol{\xi}_{1},\dots,\boldsymbol{\xi}_{m}$ on $\partial{\bar{U}}$ where each $\boldsymbol{\xi}_{k}=(\xi_{k}^{1},\dots,\xi_{k}^{N})\in\mathbb{C}^{{N}}$. 
For $\mathbf{x}=(x_{1},\ldots,x_{N})\in M$, let us define
\[
	K\left(\boldsymbol{\xi}_{k},\mathbf{x}\right):=\prod_{i=1}^{N}{\frac{1}{\xi_{k}^{i}-x_{i}}}.
\]
\end{definition}

Since ${M}$ is compact, and lies strictly within ${{U}}$,
every denominator $\xi_{k}^{i}-x_{i}$ is bounded away from zero, ensuring that $K(\boldsymbol{\xi}_{k},\mathbf{x})$ is uniformly bounded. Moreover, small variations in $\mathbf{x}$ produce smooth changes in $K(\boldsymbol{\xi}_{k},\mathbf{x})$, making the kernel a robust tool for approximating functions with steep gradients or near-singularities. Notably, the inversion-based form of the kernel is directly related to the activation function used in CauchyNet \textcolor{black}{(Sec. 3)}.

\begin{figure}[H]
    \centering
    \includegraphics[scale=0.105]{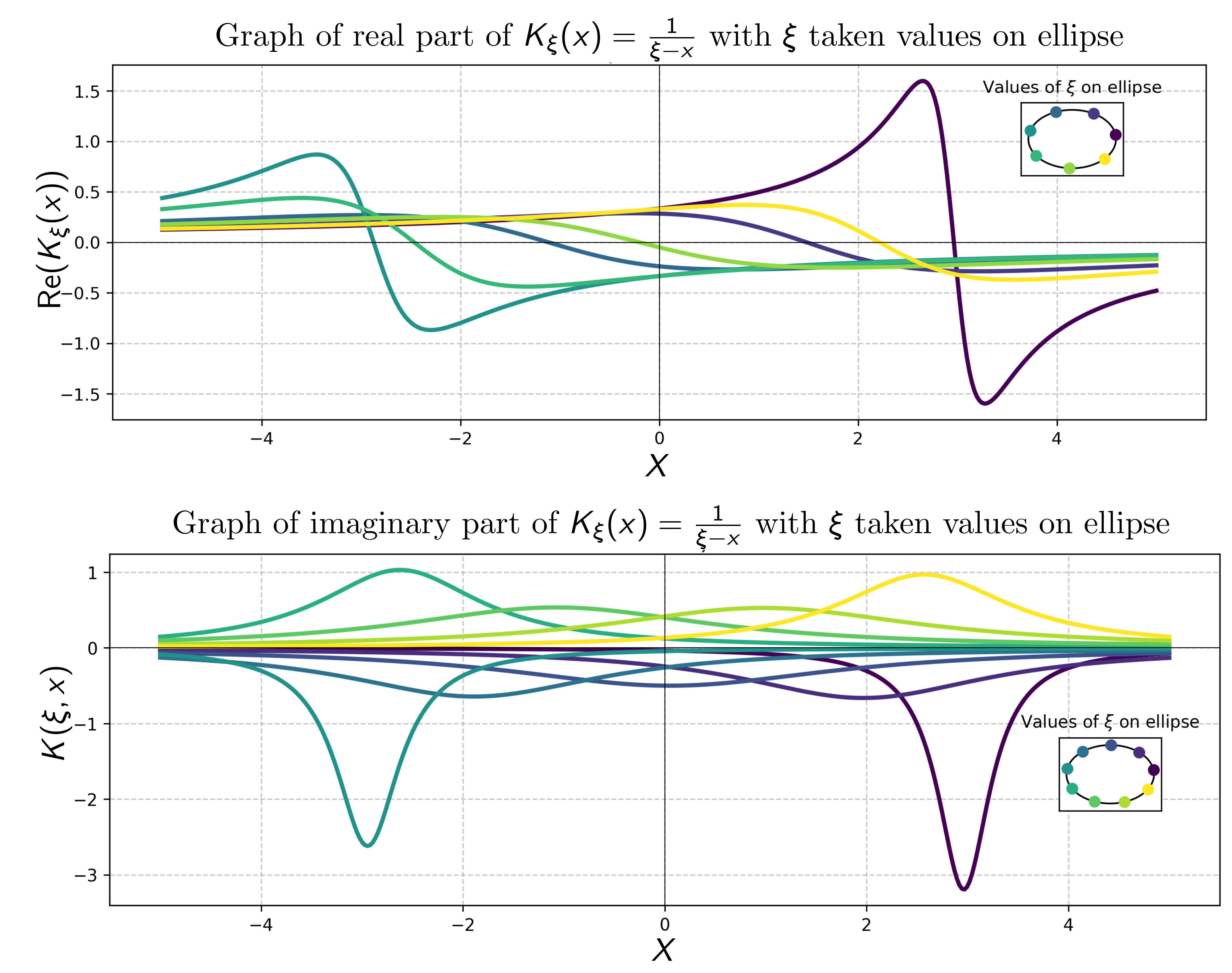}
    \caption{Behavior of the one-dimensional complex Cauchy kernel $K(\boldsymbol{\xi}, x) = 1$, with $\boldsymbol{\xi}$ values distributed along an ellipse in the complex plane. The main plots show the real (top) and imaginary (bottom) parts of $K(\boldsymbol{\xi}, x)$ as a function of $x\in[5, 5]$. The inset illustrates the positions of $\boldsymbol{\xi}$ values on the ellipse, with semi-major and semi-minor axes of 6 and 2 units, respectively. This visualization highlights the kernel's ability to encode variations in $x$ while preserving holomorphic properties.}
    \label{1DCauchyKernel}
\end{figure}

Fig.~\ref{1DCauchyKernel} illustrates the behavior of the real and imaginary parts of $K(\boldsymbol{\xi},\cdot)$, for different $\boldsymbol{\xi}$ values along an ellipse in the complex plane.

\subsection{Universal Approximation for CauchyNet}\label{UniApprox4CN}
A central insight is that linear combinations of Cauchy kernels can approximate any continuous function on $M$ to arbitrary accuracy. We formalize this in the following theorem.

\begin{theorem}{\rm(Cauchy Approximation Theorem).}\label{Thm1}
Let $f\in{\mathcal{F}}_{M}$ be continous on $M\subset\mathbb{R}^{N}$ with $M$ contained in an open domain $U\subset\mathbb{C}^{N}$ whose boundary is $\partial \bar{U}$. For any $\epsilon>0$, 
there exists $m$, points $\{\boldsymbol{\xi}_{1},...,\boldsymbol{\xi}_{m}\}\subset\partial\overline{{{U}}},$ and complex
coefficients $\{\theta_{1},...,\theta_{m}\}\subset\mathbb{C}$ such that
\[
	\operatorname*{sup}_{\mathbf{x}\in M}{\Big|}f(\mathbf{x})-\sum_{k=1}^{m}\theta_{k}\,K\big(\boldsymbol{\xi}_{k},\mathbf{x}{\big)}{\Big|}\,<\epsilon.
\]
\end{theorem}

This theorem guarantees that by selecting a sufficient number of points on
${\partial \bar{U}}$ and appropriate weights $\theta_{k}$, the kernel
expansion can uniformly approximate any continuous function on $M$
(A detailed proof is provided in the {supplement}).  
Furthermore, we have: 
\begin{theorem}
The Cauchy kernel function is dense in the function space ${\mathcal{F}}_{M}.$
\end{theorem}

Because CauchyNet learns complex biases analogous to the $\boldsymbol{\xi}_k$ and an output vector analogous to $\theta_{k}$, it naturally inherits
the following universal approximation property. Let $C_{\mathrm{net}}$ denote
the collection of all CauchyNets.
\begin{theorem}{\rm(Universal Approximation Theorem for CauchyNet).}\label{ThmUniApprox}
Let $f:M\to\mathbb{R}$ be a continuous function defined on a compact domain $M \subset \mathbb{R}^{N}$ with
$M\subset U\subset{\mathbb{C}}^{N}$. 
For any $\varepsilon>0$, there exists an integer $h\geq1$, and a CauchyNet  
$N_{\theta,h}\in{\cal C}_{net},$ parameterized by $\boldsymbol{\theta}=(\mathbf{B},\mathbf{c})$, with
hidden layer width ${h},$ such that
\[
	\left|f(\mathbf{x})-N_{\boldsymbol{\theta},h}(\mathbf{x})\right|<\varepsilon,
\]
for all ${\bf x}=(x_{1},x_{2},...,x_{N})^{\mathrm{T}}\in{M}.$ 
\end{theorem}

Proof. Each row
$\mathbf{b}_{k}$ of ${\bf B}\in{\mathbb C}^{h\times N}$ serves as a point
$\boldsymbol{\xi}_k$ on $\partial \bar{U}$, and the inversion-based activation
\[
	\prod_{i=1}^{N}(b_{k,i}-x_{i})^{-1}
\]
corresponds to the Cauchy kernel $K(\xi_{k},\mathbf{x})$. By choosing the
output coefficients in $\mathbf{c}$ to match the weights $\theta_k$,  the linear combination produced by CauchyNet approximates $f(\mathbf{x})$ arbitrarily closely, as guaranteed by {Theorem}~\ref{Thm1}.
$\hfill\blacksquare$

Theorem~\ref{ThmUniApprox} shows that for any continuous function
${f}$ and any $\varepsilon>0$, a sufficiently large CauchyNet
exists within $C_{\mathrm{net}}$ to approximate $f$ to within
$\varepsilon$.
In summary, CauchyNet requires only about $2\,h\left(
m+1\right)$ real parameters, and has computational complexity ${\ {\ {\mathcal{O}}}}(h\,m)$, making it efficient for real-time applications on resource-constrained hardware.
Its \emph{holomorphic}, inversion-based activation ensures robust gradient
flow, which mitigates common issues such as vanishing or exploding gradients. Although CauchyNet is primarily designed for function approximation in data-scarce settings, its framework can be adapted for tasks like time-series forecasting and missing-value imputation. We next validate these theoretical findings with numerical experiments{(Sec. 6 and the supplement)}.

\begin{figure*}[ht]
    \centering
    \includegraphics[scale=0.085]{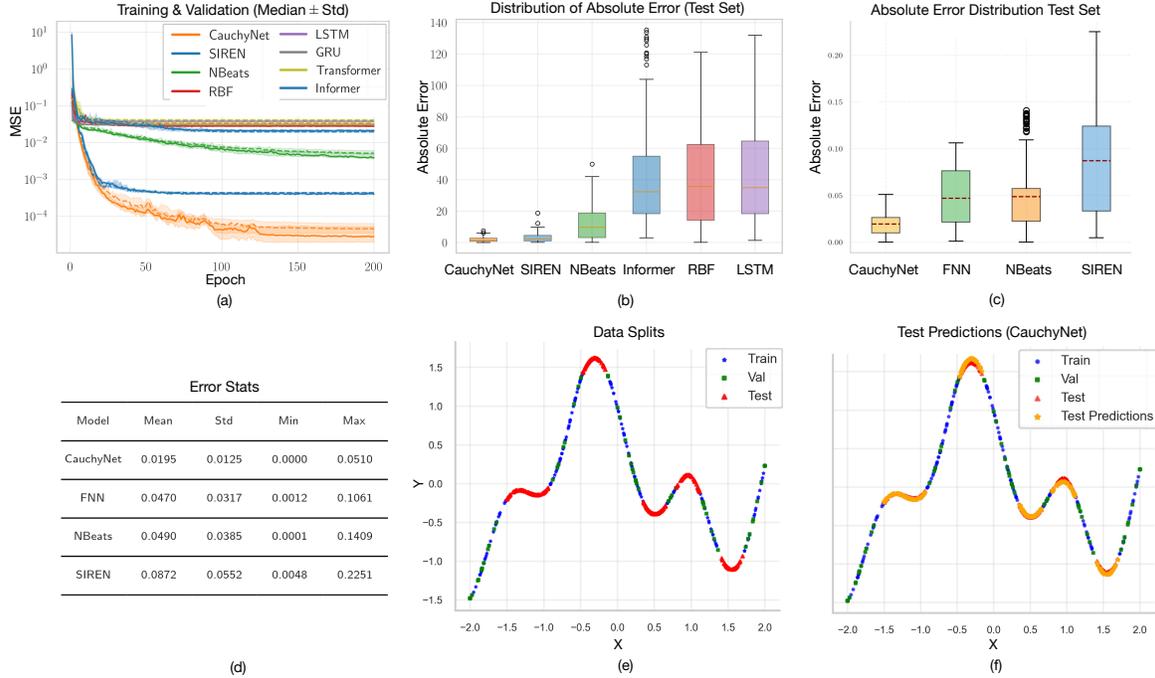}
    \caption{(a) Training and validation loss (log scale) over 200 epochs. CauchyNet converges faster and retains lower validation loss. (b) Box plot of absolute errors on the test set for various models. CauchyNet shows the smallest median error and least variability, excelling under data scarcity (MAE of 1.5 vs. 3.2 or much higher for baselines). (c) Box plot of absolute errors on the 1D test set for various models. (d) A table listing of error statistics of CauchyNet and other baseline models. CauchyNet achieves the smallest median error and the tightest error distribution, significantly outperforming baselines like SIREN, N-BEATS, and FNN in imputing missing values. (e) Training (blue) and test (red) points for the one-dimensional gap-filling task, with missing zones centered around turning points. (f) Predictions from CauchyNet (orange) closely match the true function (dashed line) within the missing regions, demonstrating accurate reconstruction.}
    \label{Exp}
\end{figure*}

\section{Experiments}\label{section_experiment}
The efficacy of CauchyNet is evaluated under real-world constraints, focusing on data-scarce scenarios and limited computational resources. The objective is to assess whether CauchyNet can achieve stable and efficient learning in demanding environments in which traditional models may falter. The experimental setup and more experimental results are provided in the supplement.

\subsection*{Experiment 1: Function Approximation with Sharp Peaks under Limited Data.} We first assess CauchyNet's ability to approximate complex functions featuring sharp peaks and near-singularities under minimal training data. The target function $f(x)$ on $[-1, 1]$ is:
\begin{align}
f(x)=&{\frac{1}{(x+0.6)^{2}+0.005}}-40\,e^{-2(x+0.4)^{2}}\notag\\ &+\ 50\operatorname{sign}(x)\,|\sin(3x)+0.8|^{1.5}\,\sin(10x).\notag
\end{align}
This function exhibits spikes near $x = 0.6$ and $x = 0.5$, plus a singularity at $x = 0$. To simulate
data scarcity, only 150 evenly spaced points are used for training, with 75 for validation and 75 for testing. 
We trained the compared networks for 200 epochs, 
and conducted paired t-tests comparing CauchyNet vs. baselines on MSE/MAE, all yielding $p < 0.05$. Fig.~\ref{Exp} (a) shows faster, more stable convergence for CauchyNet; (b) the boxplot reveals its lower median error and tighter distribution compared to the others.


CauchyNet achieves the lowest MSE/MAE for approximating sharp peaks and singularities, converges
more rapidly, and has minimal error variance. Its rational activation and compact design enable
stable training with fewer parameters, making it attractive for real-world tasks requiring accurate
modeling of abrupt changes.

\subsection*{Experiment 2: Imputation of Missing Values Using CauchyNet}
We test CauchyNet's capacity to impute missing data in one-dimensional settings, especially in zones with steep gradients or near-singularities.

\noindent\textbf{One-Dimensional Gap-Filling.} 
We define the target function $g(x)$ on $[-2, 2]$ combining trigonometric, polynomial, and rational components:
\begin{align}
 g(x)=&\sin(2x-4)+0.5\cos(5x-5)+{\frac{0.05}{(x-1)^{2}+0.1}}\notag\\&+\ {\frac{0.01}{(x+0.5)^{2}+0.05}}-0.01(x^{2}-x^{3}).\notag
\end{align}
This function features six turning points, see Fig.~\ref{Exp}(e), each surrounded by a ±0.15 interval where
the 150 training data is withheld, simulating missing zones. CauchyNet is trained for up to 10,000
epochs. It accurately reconstructs these missing zones (Fig.~\ref{Exp}(f). This function features six turning
points, see Fig.~\ref{Exp}, each surrounded by a ±0.15 interval where the 150 training data is withheld,
simulating missing zones.


CauchyNet accurately reconstructs the withheld regions, whereas baselines like SIREN, N-BEATS, and FNN, exhibit overshooting, underfitting, or instability. Fig.~\ref{Exp} (c) illustrates the boxplots of absolute errors, confirming CauchyNet's superior and stable performance. 
Fig.~\ref{Exp}(d) summarizes the imputation performance metrics. CauchyNet achieves the lowest MSE and MAE, confirming its effectiveness among one-dimensional imputation tasks.

\section{Conclusion and Future Work}
This work introduced CauchyNet, a compact neural network inspired by \emph{Cauchy's integral formula}, and tailored for resource-constrained, low-data environments. The evaluations on synthetic functions, M4 dataset subsets, and missing-value imputation tasks, demonstrate that CauchyNet achieves superior performance, compared with state-of-the-art methods including Transformers and N-BEATS, while requiring significantly fewer parameters. These findings underscore CauchyNet's promise for efficient computation in applications like IoT systems and real-time edge computing. 
By embedding reciprocal structures from complex analysis, CauchyNet effectively bridges classical approximation theory with modern neural network design. Its \emph{holomorphic inductive bias} and \emph{compact} architecture, ensure stable training and reliable performance across diverse tasks, including forecasting, imputation, and scientific computing. 
While the study validates CauchyNet under resource-constrained and low-data scenarios, several avenues remain to enhance its applicability and performance:\\ 
\noindent\textbf{Scalability and Generalizability}. Future research will extend evaluations to larger and more diverse datasets (e.g., the full M4 benchmark and high-dimensional time series), to thoroughly assess scalability and generalization. Although the experiments focus on low-dimensional tasks, future work will explore high-dimensional inputs and longer sequences. While the architecture's inherent efficiency $(\mathcal{O}(hm))$ is promising, detailed studies on scalability remain essential. 

\noindent\textbf{Architectural Enhancements}. The current theorectical analysis focuses on single-layer models. We plan to explore deeper and residual variants, to boost expressivity and capture complex dependencies, ensuring that such modifications maintain optimization stability and computational efficiency. 

\noindent\textbf{Extended Applications}. To broaden the practical utility of CauchyNet, we will investigate: 1) New regularization techniques tailored for complex-valued parameters, to improve robustness in noisy or data-scarce environments. 2) Incremental or online learning frameworks to enable real-time data handling in sensor networks and IoT systems. 3) Integration with broader computational frameworks, such as partial differential equation solvers and neural operators, to model steep gradients and localized singularities in scientific simulations. These directions will help further validate CauchyNet's efficiency, and extend its applicability across a wider range of challenging, real-world tasks including generative models~\citep{song2021scorebasedgenerativemodelingstochastic,10.5555/3495724.3496298,WGAN,lou2024discrete} for high-resolution images, videos, and textual data.


\bibliographystyle{abbrvnat}
\bibliography{main}

\onecolumn

\section{Common Neural Architectures and Their Characteristics}
Fig.~\ref{common_architecture} presents a comparative snapshot of prevalent neural architectures, highlighting parameter overhead, data efficiency, and typical limitations. Even with pruning or compression (Han et al., 2015; Iandola et al., 2016), many models require large datasets and tend to overfit when data are scarce.

\section{Notaitons and Symbols}
Table~\ref{table:notation} presents the key mathematical notations, symbols and the detailed descritions.
\begin{table*}[ht]
\centering
\caption{Key Mathematical Notaitons and Symbols.}\label{table:notation}
\begin{tabular}{lll}
\hline
\textbf{Symbol} & \textbf{Definition} & \textbf{Usage} \\
\hline
${M}$ & $M=\prod_{i=1}^{N}M_{i}\subset\mathbb{R}^{N}$ &
{The compact domain where the target function is defined.}\\
$U$ & $U=\prod_{i=1}^{N}U_{i}\subset\mathbb{C}^{N}$ &  \makecell{A complex domain containing $M$, used for holomorphic extension.} \\
$\bar{{U}}$ & Closure
of ${\big.}U{\big.}{\big.}$ & The closed set of ${U}$, with boundary
$\partial{{\bar{U}}}$. \\
${\mathcal{F}}_{M,D}$ & $C^{0}(M,D)$ & The space of continuous functions on ${{M}}$.\\
$f(\mathbf{x})$ & Target function on ${M}$ &  The function we wish to
approximate.\\
$\mathbf{x}$ & Input vector in $\mathbb{R}^{N}$ & A
point in $M$.\\
${\boldsymbol{\xi}}_{k}$ & Point on $\partial{\bar{U}}$ & A sample from the boundary of ${\mathbf{}}U$, used
in the kernel.\\
 $K(\boldsymbol{\xi}_{k},\mathbf{x})$ & $\prod_{i=1}^{N}{\frac{1}{\xi_{k}^{i}-x_{i}}}$ &  The Cauchy kernel function.\\
 ${\mathcal{\theta}}_{k}$ & Complex coefficient & Weight for the $k$-th kernel in the approximation.\\
 $\epsilon$ & Error tolerance & The desired approximation accuracy.\\
 $N_{\theta,h}$ & CauchyNet approximator & 
The network with parameters $\boldsymbol{\theta}$ and ${h}$
hidden neurons.\\
$h$ & Hidden layer width & The number of hidden neurons in CauchyNet.\\
$\mathbf{B}$ & Bias matrix in $\mathbb{C}^{h\times N}$  & Learnable biases for the hidden neurons.\\
$\mathbf{c}$ &  Coefficient vector in $\mathbb{C}^{h}$ & Weights for combining hidden outputs.\\
$\varepsilon$ &
Activation offset & A small constant added to avoid division by zero.\\
\hline
\end{tabular}
\end{table*}

\begin{figure}
    \centering
    \begin{framed}
\begin{itemize}
  \item LSTM~\citep{10.1162/neco.1997.9.8.1735}
  \item[] \textbf{Strengths}: Effectively learns long-term temporal dependencies via gating.
  \item[] \textbf{Weaknesses}: Large parameter footprint and sensitive hyperparameters, especially in sparsedata
contexts.
\item Transformer~\citep{NIPS2017_3f5ee243} 
\item[] \textbf{Strengths}: Captures long-range correlations efficiently with abundant data.
\item[] \textbf{Weaknesses}: Quadratic complexity in sequence length; memory-intensive.
\item Informer~\citep{zhou2021informerefficienttransformerlong}
\item[] \textbf{Strengths}: Employs sparse self-attention to enhance scalability for long sequences.
\item[] \textbf{Weaknesses}: Multi-layer stacks involve thousands of parameters and require large datasets.
\item N-BEATS~\citep{oreshkin2020nbeatsneuralbasisexpansion}
\item[] \textbf{Strengths}: Provides state-of-the-art forecasting accuracy with interpretable decompositions
into trend and seasonal components.
\item[] \textbf{Weaknesses}: High parameter count and data-hungry; may overfit in scenarios with limited
data.
\item MLP
\item[] \textbf{Strengths}: Simple and straightforward to implement and train.
\item[] \textbf{Weaknesses}: Lacks specialized inductive biases for oscillatory or near-singular data.
\item RBF Network~\citep{ParkSandberg1991}
\item[] \textbf{Strengths}: Excels at local approximations and can model a broad class of functions.
\item[] \textbf{Weaknesses}: Placement of kernel centers is non-trivial; struggles with sharp rational spikes
or steep gradients.
\item SIREN~\citep{NEURIPS2020_53c04118}
\item[] \textbf{Strengths}: Sinusoidal activations capture smooth and high-frequency signals effectively.
\item[] \textbf{Weaknesses}: Requires large capacity to model abrupt spikes or extremely limited data; not
explicitly designed for missing-data imputation.
\item TCN~\citep{bai2018empiricalevaluationgenericconvolutional}
\item[] \textbf{Strengths}: Utilizes dilated convolutions for efficient sequence modeling.
\item[] \textbf{Weaknesses}: Deeper architectures increase parameter counts and can be computationally
slow.
\end{itemize}
\end{framed}
\caption{Common Neural Architectures and Their Characteristics}
\label{common_architecture}
\end{figure}

\section{Experiments}

This section details the experimental setup, and provides more experimental results.

\noindent\textbf{Common Setup and Metrics.} 
All the experiments follow a standardized training procedure to ensure consistency. Models are evaluated using Mean Squared Error (MSE), Mean Absolute Error (MAE), training time and number of parameters. The primary training configurations are given in Table~\ref{table:traning_setup}.

\begin{table*}[ht]
\centering
\caption{Common Training Setup Across All Experiments.}\label{table:traning_setup}
\begin{tabular}{ll}
\hline
\textbf{Parameter} & \textbf{Value} \\
\hline
Data Split Ratio & 50\% training, 25\% validation, 25\% testing \\
Architecture & One hidden layer with 128 neurons \\
Batch Size & 32 \\
Optimizer & Adam, learning rate 0.01, decayed by factor 0.5 every 100 epochs \\
Weight Decay & $1 \times 10^{-4}$ \\
Preprocessing & Target values normalized via \texttt{MinMaxScaler} \\
Random Seed & Fixed at 10 for reproducibility \\
Imaginary-Part Penalty & $\lambda = 0.1$ (unless stated otherwise) \\
\hline
\end{tabular}
\end{table*}

All models use a single hidden layer with 128 neurons, though the total parameter count differs
by architecture. As shown in Table~\ref{table:parameters}, CauchyNet maintains a significantly lower parameter count compared to other baselines, highlighting its efficiency.

\begin{table*}[ht]
\centering
\caption{Number of Parameters for Each Model.}\label{table:parameters}
\begin{tabular}{lccccccc}
\hline
Model & CauchyNet & SIREN & N-BEATS & Informer & Transformer & RBF & LSTM \\
\hline
\#Params & 256 & 385 & 385 & 149{,}377 & 132{,}993 & 385 & 6{,}720 \\
\hline
\end{tabular}
\end{table*}

\noindent\textbf{Hardware Configuration and Resource Usage.} 
All experiments were conducted on a Mac
with an Apple M3 chip, using its integrated GPU. Even for larger-scale tasks (e.g., M4 subset, 2D
polynomial-rational surface), GPU memory usage remained within 2 GB, demonstrating efficiency
on resource-limited hardware. Training times for CauchyNet (128 hidden neurons) ranged from $\sim30$ s for simpler synthetic datasets to $\sim2$ min for 2D surface extrapolation with 3,000 samples,
reinforcing its suitability for edge computing and low-power environments.

\noindent\textbf{Scalability with Input Size.} Although most experiments here focus on 1D or 2D inputs, the
observed training times align with theoretical $\mathcal{O}(hm)$ complexity, showing efficiency for moderate
hidden dimensions. For instance, training on a 2D polynomial-rational surface with 3,000 samples took about 2 min per 1,000 epochs. Future studies will extend to higher-dimensional inputs and
larger real-world datasets.

\subsection*{Experiment 2: Imputation of Missing Values Using CauchyNet}

\begin{figure}[H]
    \centering
    \includegraphics[scale=0.07]{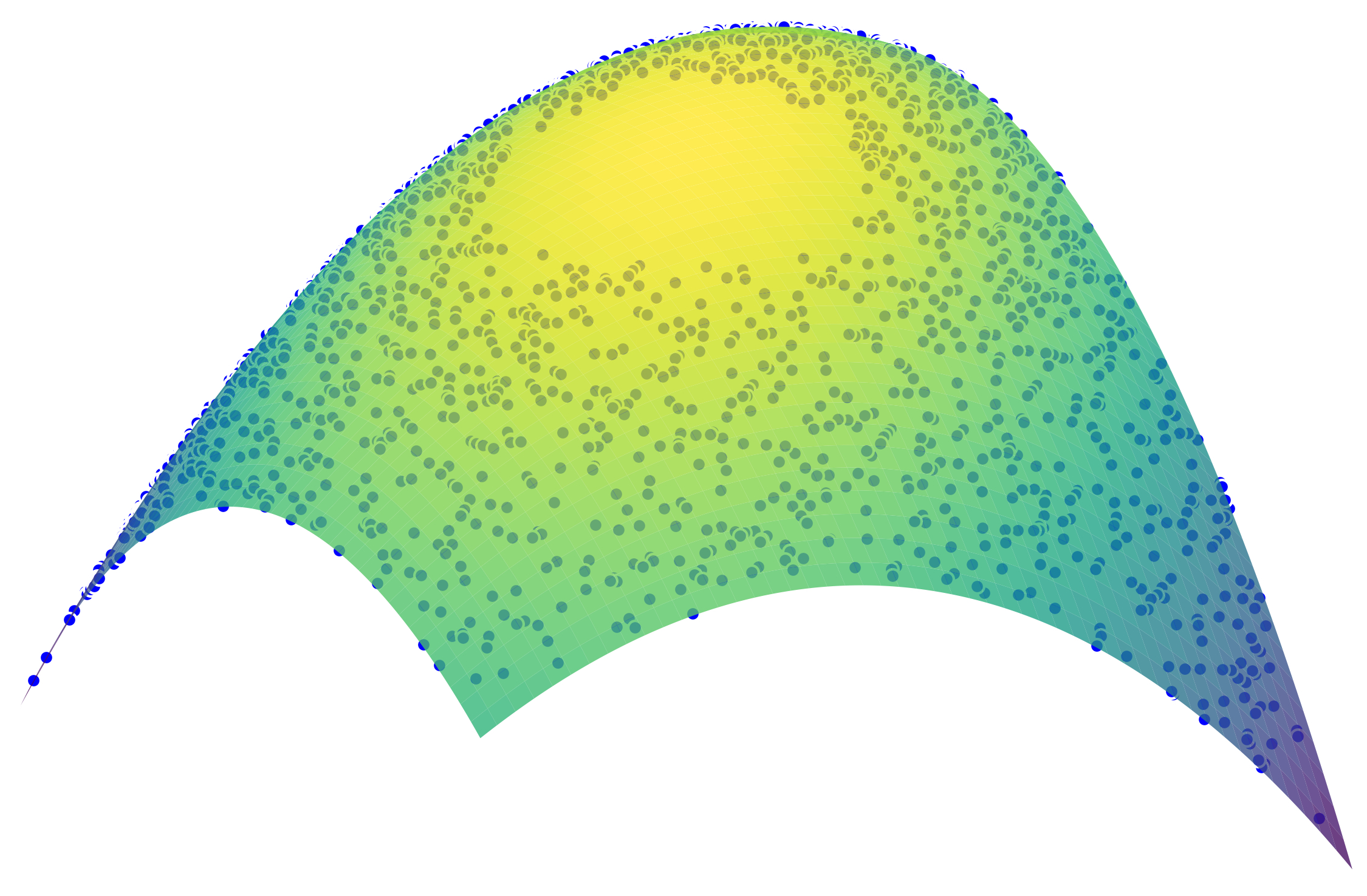}
    \hspace{5em}
    \includegraphics[scale=0.08]{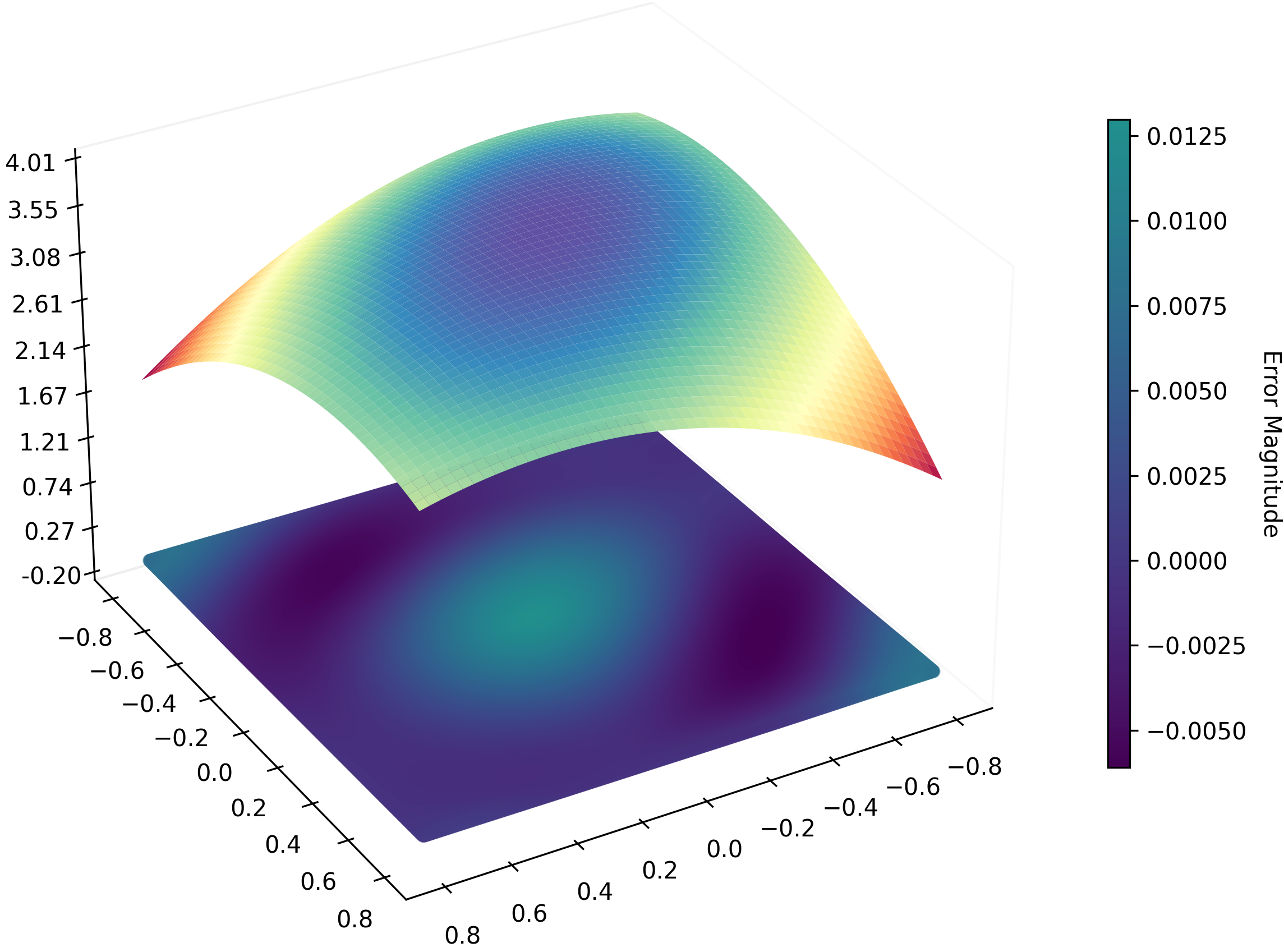}
    \caption{(Left) Training points (blue) exclude the circular radius 0.3 around (0, 0) (red test points). The polynomial-rational function $g(x, y)$ can approach singularities where $(x-1)^2$ is small. (Right) Top: Surface predicted by CauchyNet over $[-0.8, 0.8]^2$. Bottom: Signed error from about -0.005 (blue) to 0.0125 (red). Despite having no training data in the central disk, CauchyNet accurately fills the gap, indicating strong extrapolation in polynomial-rational domains.}
\label{fig_training_pts_circular}
\end{figure}

\noindent\textbf{Two-Dimensional Polynomial-Rational Surface Imputation.}
To extend our analysis, we evaluate CauchyNet on a 2D surface with a deliberately excluded circular
region, simulating missing data in high-dimensional contexts. 
We define the target surface $g(x, y)$ over $[-0.8, 0.8]^2$ as:
\begin{align}
g(x,y)=3-x^{2}+x y-y^{2}-{\frac{1}{5+(x-1)^{2}}}.\notag
\end{align}
A circular region with radius 0.3 around the origin is excluded from training, forcing models to
extrapolate within this "missing disk".
The training setup for this experiment follows the same configurations as in Experiment 1,
except that we randomly sample 3,000 points in the domain $[-0.8, 0.8]^2$ and mark those within
radius 0.3 of (0, 0) as test, forming a "missing disk." All remaining points form train and validation
sets (in about a 60/40 ratio). This ensures the model sees no direct samples near (0, 0), forcing it to
extrapolate across potential near-singular effects.  
Fig.~\ref{fig_training_pts_circular} (left) shows the withheld disk (red). Despite no training data there, CauchyNet accurately
reconstructs the missing region (Fig.~\ref{fig_training_pts_circular} (right)), with error bounded in $[-0.005, 0.0125]$. Baselines incur larger errors, showing less effective extrapolation into near-singular areas.

CauchyNet's reciprocal activation and rational-based form allow robust extrapolation into missing or high-curvature domains, essential for sensor grids, imaging, and other real-world data imputation tasks.


\subsection*{Experiment 3: Approximating a Two-Dimensional Polynomial-Rational Surface.} 
While many neural networks excel in one-dimensional function approximation or time-series tasks,
higher-dimensional problems often pose greater challenges, especially when surfaces exhibit mixed
polynomial-rational components. Here, we evaluate CauchyNet's ability to learn a nontrivial twodimensional
(2D) surface under limited data and localized complexity. Success in this domain highlights the model's broader applicability to phenomena that cannot be decomposed into simple one-dimensional patterns (e.g., partial differential equations, multi-sensor grids, or manufacturing
processes). 
We next test CauchyNet on a 2D domain to confirm its broader applicability. The target surface
$g(x, y)$ over $[-1.5, 1.5]^2$ is:
\begin{align}
g(x,y)=x^{2}-x y+3y+y^{2}+{\frac{1}{5+x^{2}}}.\notag
\end{align}
We randomly sample 300 points, splitting them 50/25/25 for train/val/test. CauchyNet is instantiated with a single hidden layer of 128 units and trained for up to 500 epochs. As in Experiment 6.2, we use Adam and MSE loss. The same random seed is applied for reproducibility, and no additional regularization is introduced to maintain fairness.

\begin{figure}[H]
    \centering
    \includegraphics[scale=0.14]{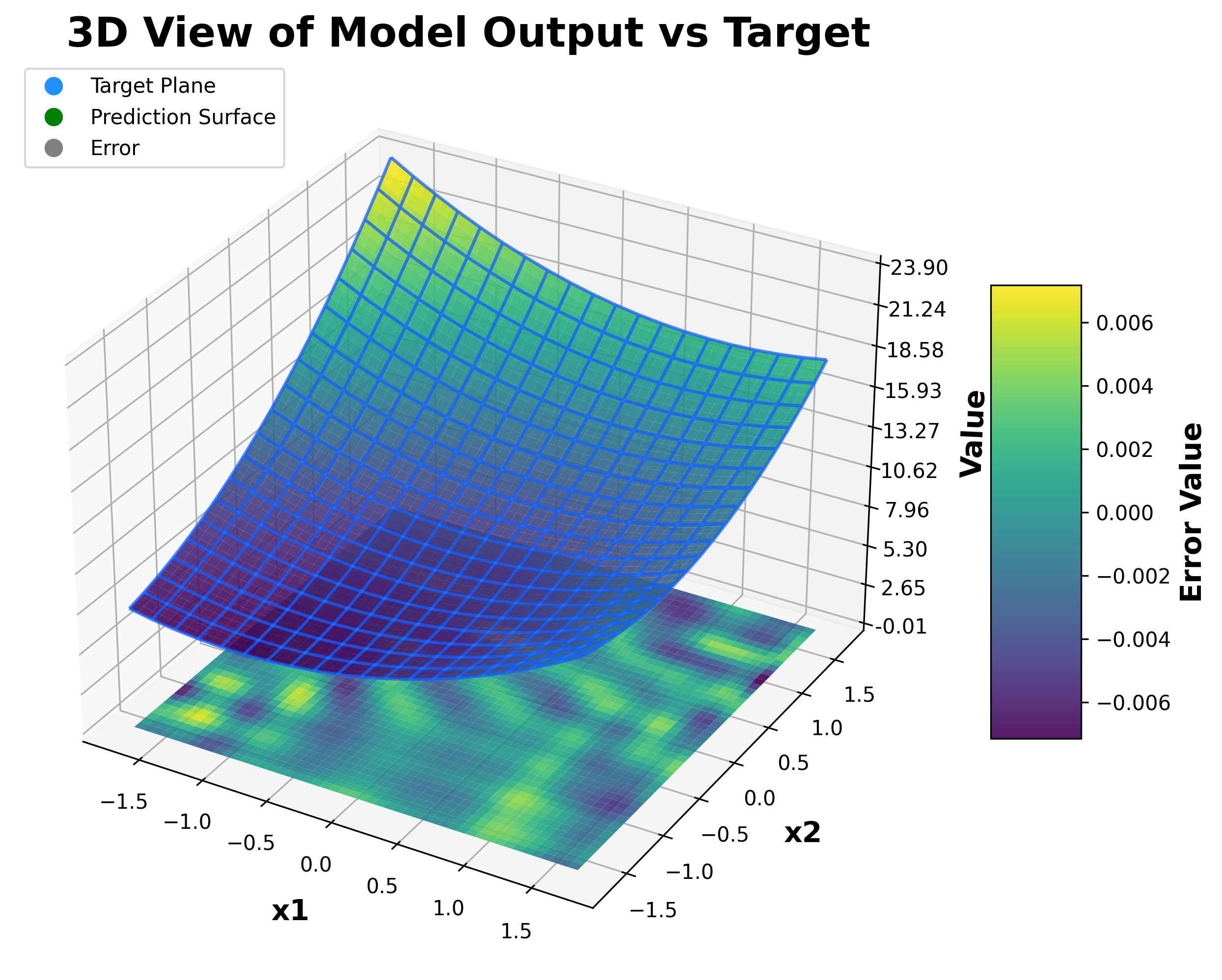}
    \caption{(Top) Predicted 2D surface by CauchyNet over $[-1.5, 1.5]^2$, approximating a target polynomial rational function. (Bottom) Error map showing discrepancies between the predicted surface and ground truth, ranging from -0.006 (blue) to 0.006 (yellow).}
    \label{fig_surface}
\end{figure}

Fig.~\ref{fig_surface} shows CauchyNet's predicted 2D surface, consisting of two main panels. The top pane displays the surface predicted by CauchyNet over the domain $[-1.5, 1.5]^2$, accurately capturing the complex polynomial-rational structure of the target function. The bottom panel shows the signed error map illustrating the discrepancies between the predicted surface and the ground truth function, ranging from $-0.006$ (blue) to $0.006$ (yellow). This result underscores the network's ability to interpolate across polynomial and rational components smoothly.

CauchyNet adapts well to 2D polynomial-rational surfaces. Its rational-based representation captures high-gradient and near-singular regions with far fewer parameters than large baselines, highlighting its efficiency and precision.

\subsection*{Experiment 4: Forecasting on the M4 Dataset under Data Scarcity}
We evaluate CauchyNet's forecasting performance on a subset of the M4 benchmark (m4c, 2018),
focusing on trend components under limited data. 

\noindent\textbf{Data Processing.} 
In addition to the synthetic dataset from Experiment 1, we utilize a subset of the M4 forecasting benchmark (m4c, 2018), comprising 700 time series from various domains (e.g., finance, demography).
Inspired by the N-BEATS architecture, we focus solely on predicting the trend component to ensure fair comparisons.

\begin{figure}[H]
    \centering
    \includegraphics[scale=0.1]{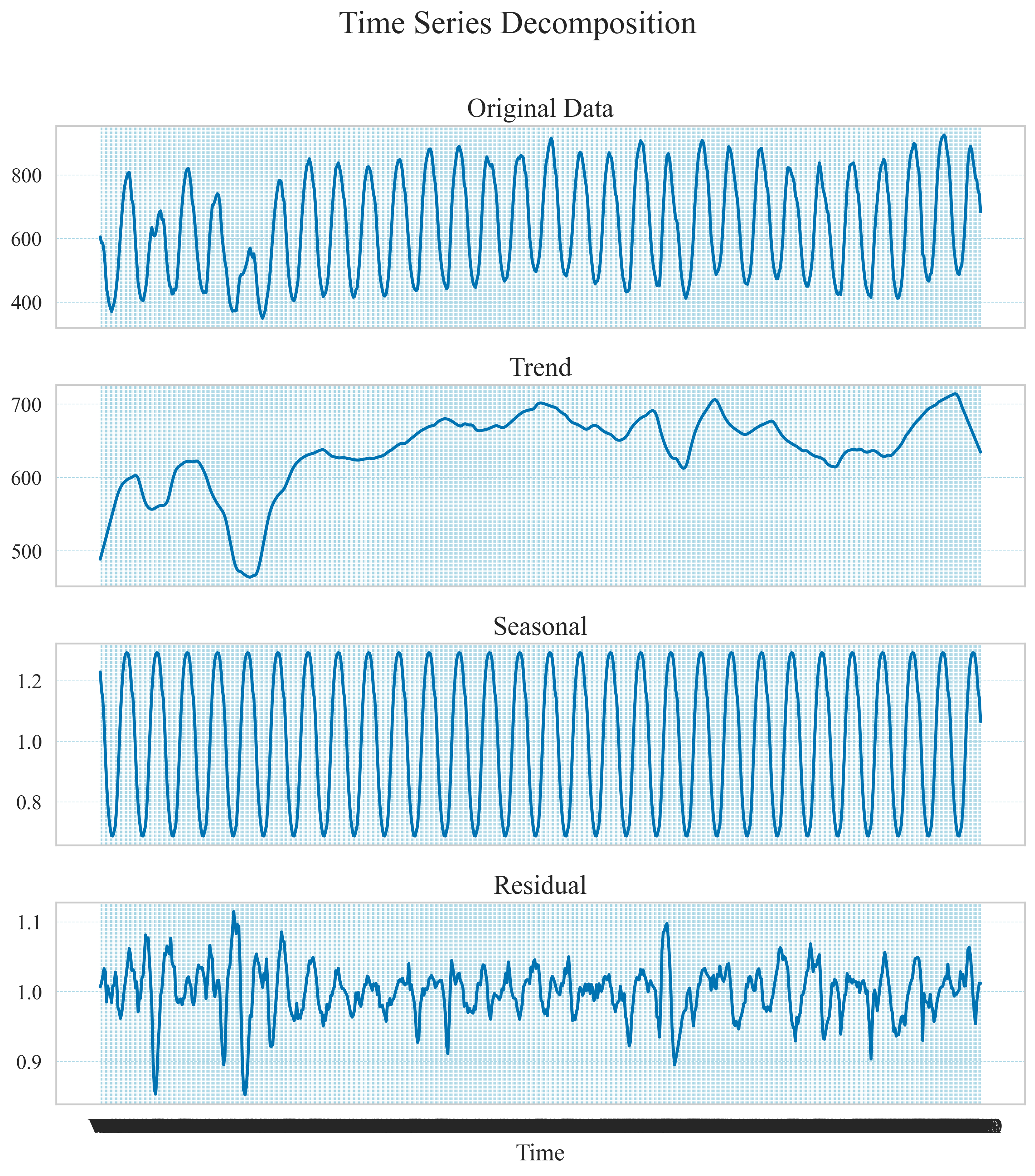}
    \caption{Time series decomposition using a multiplicative model, separating the original time series into
trend, seasonal, and residual components. This preprocessing step is used in Experiment 6.3 to focus on
predicting the trend component, which captures the long-term patterns of the data.}
    \label{fig_time_series_decompose}
\end{figure}

\begin{figure}[H]
    \centering
    \includegraphics[scale=0.12]{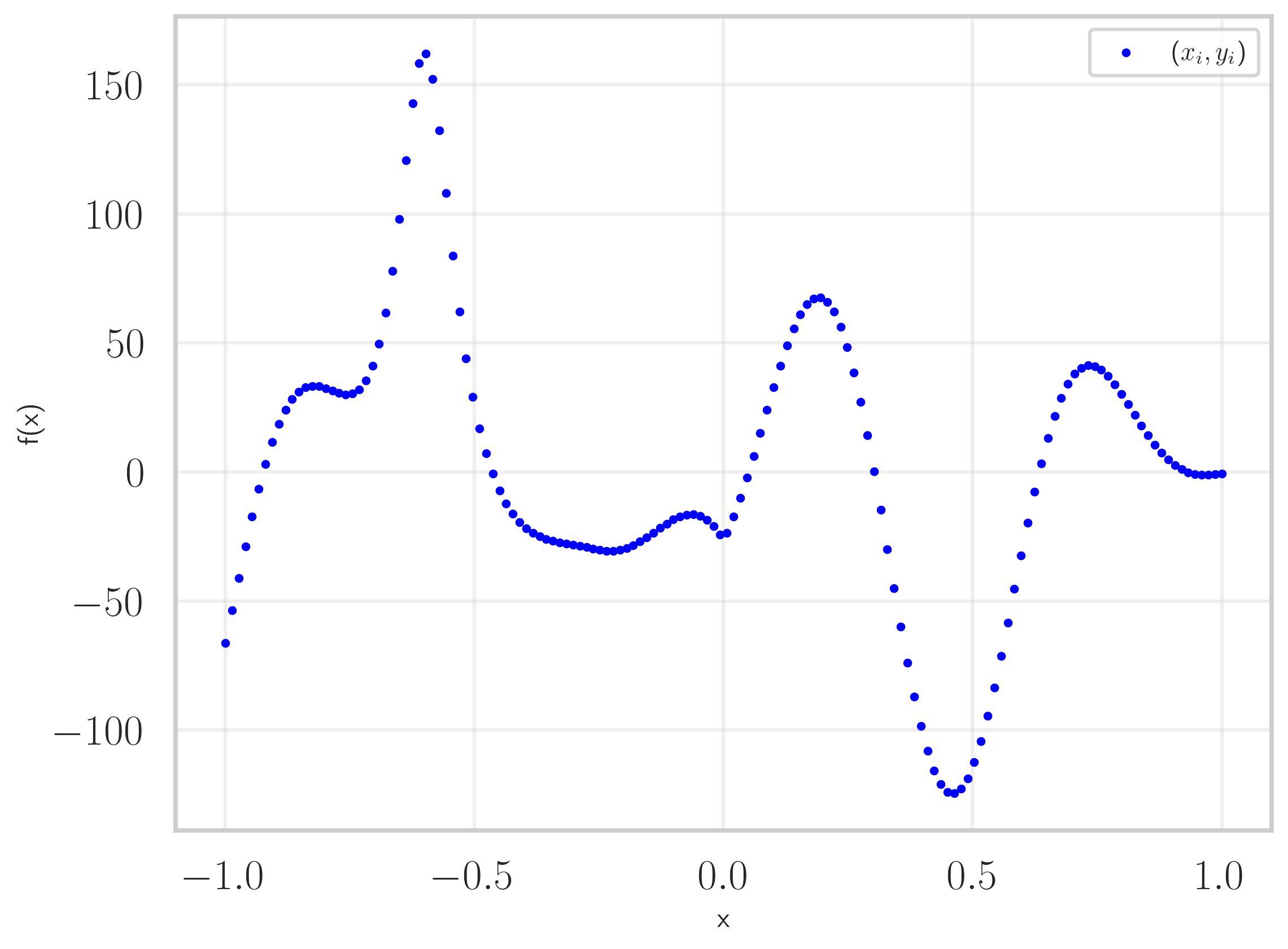}
    \includegraphics[scale=0.12]{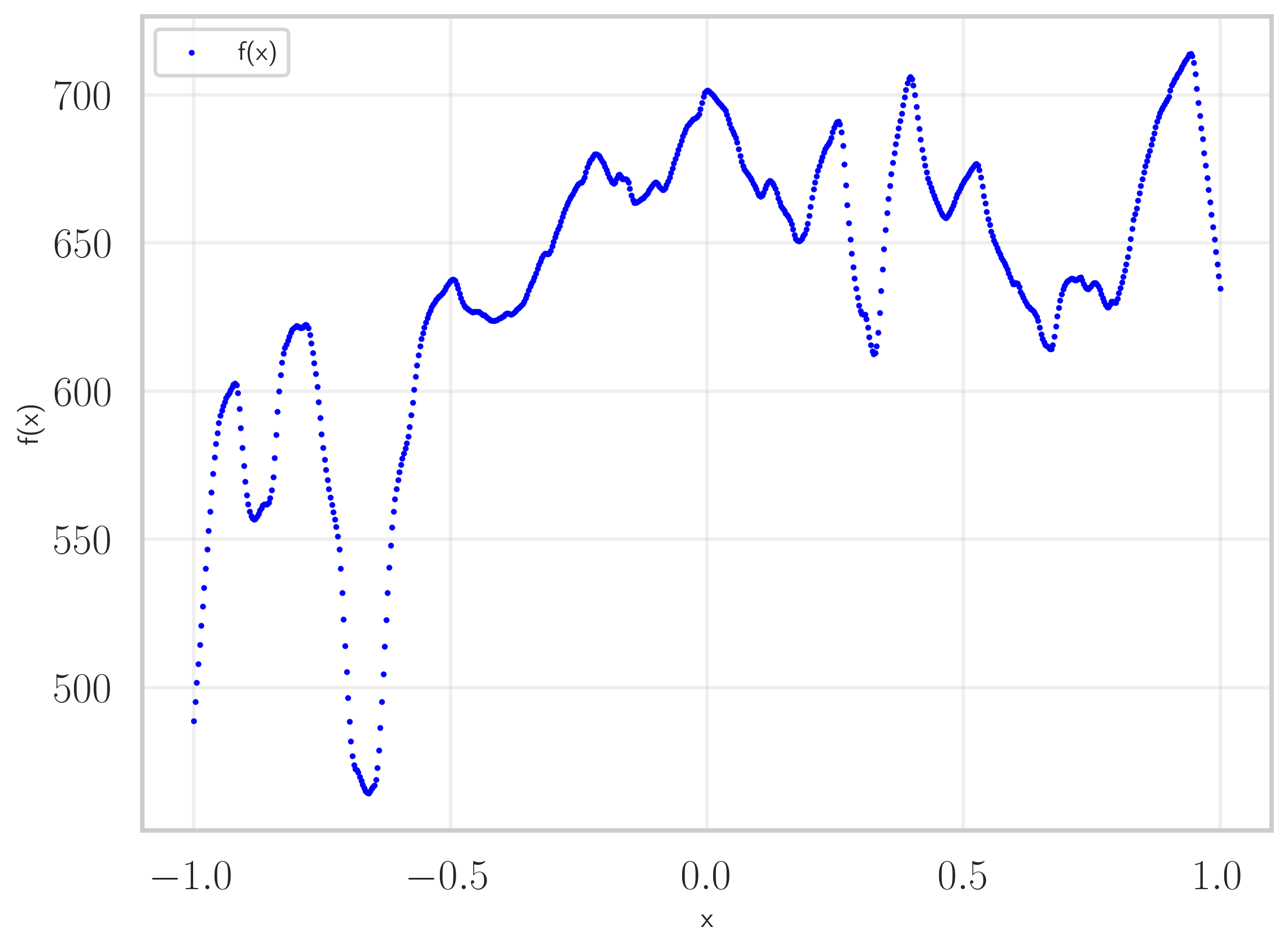}
    \caption{(Left) Target function with sharp peaks and singularities. Only 150 points (blue) are used for
training. (Right) Sample time series trend data from M4 used in Experiment 6.4, illustrating variety in
datasets.}
    \label{fig_target_function}
\end{figure}

Each time series undergoes seasonal decomposition using a multiplicative model, separating it into trend, seasonal, and residual components (see Figure~\ref{fig_time_series_decompose}). The seasonal component, proportional
to the series level and exhibiting consistent periodic patterns, enhances forecast accuracy without
requiring direct prediction. The residuals consist of noise and irregular variations, which are excluded
from the prediction focus. Consequently, only the trend component is normalized to [-1, 1] and used for training and evaluation, emphasizing underlying drift patterns. We utilize a subset of the M4 dataset, comprising a time series with 700 data. Despite being larger than that of Experiment 1, this subset remains limited for diverse real-world data, testing models'
generalization in complex settings. The trend component of the time series is normalized to [-1, 1] and used for training and evaluation to focus on underlying drift patterns. The training setup for Experiment 2 follows the common configurations outlined in Table 2, with 350 training/175 validation/175 test data. Figure~\ref{fig_target_function} (right) shows the dataset of this experiment.

\clearpage

\noindent\textbf{Results.} 
Fig. 6 presents two key metrics from Experiment 4: the training and validation loss trajectories
over 200 epochs (left subplot) and the box plot of absolute errors on the test set for each model (right subplot). CauchyNet demonstrates faster convergence and consistently lower validation loss
compared to baselines, as well as the smallest median error and tightest error distribution. Paired
t-tests confirm statistical significance $(p < 0.05)$.

\begin{figure}[H]\label{fig_training_validation_loss_for200}
    \centering
    \includegraphics[scale=0.12]{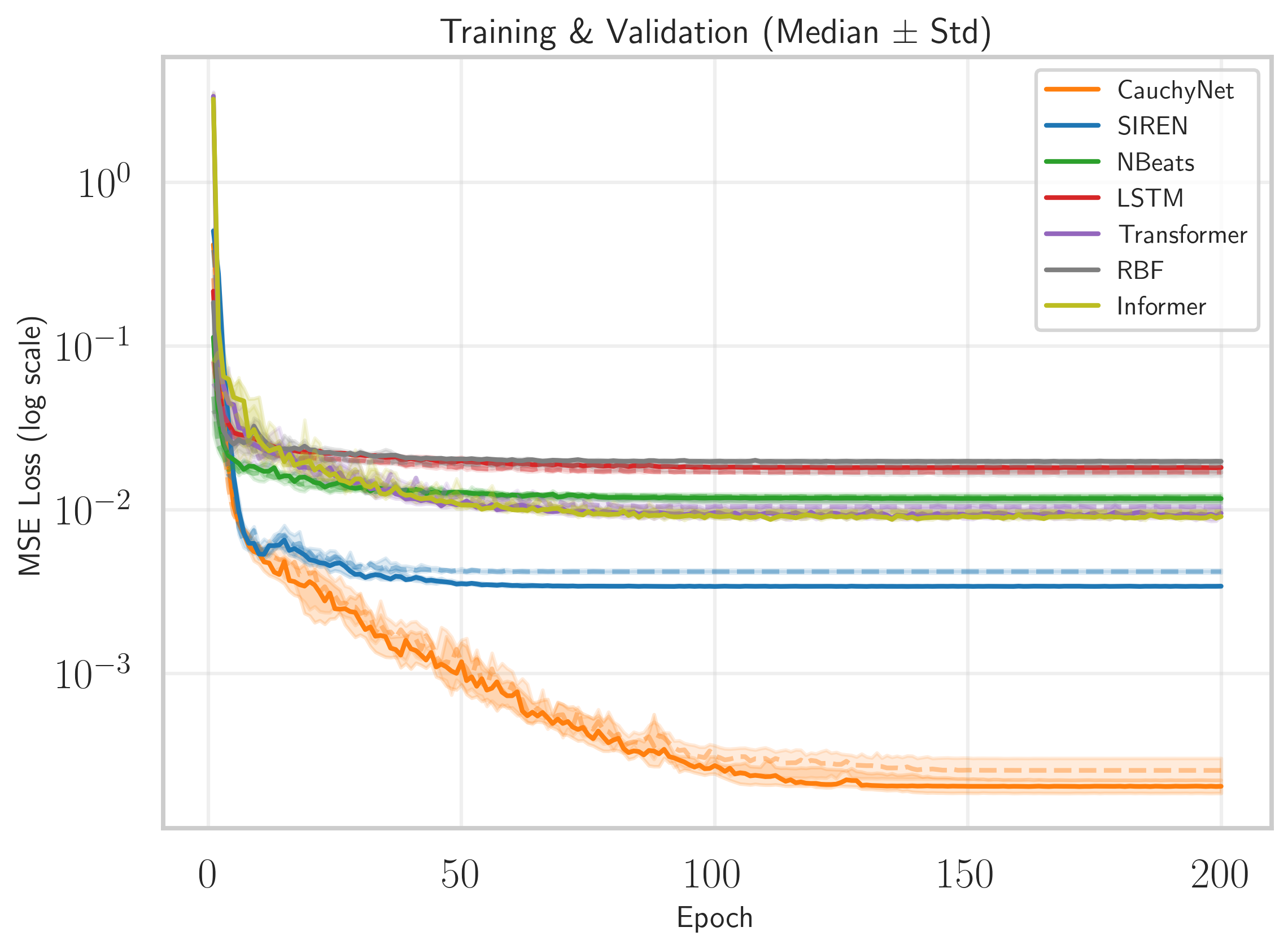}
    \includegraphics[scale=0.98]{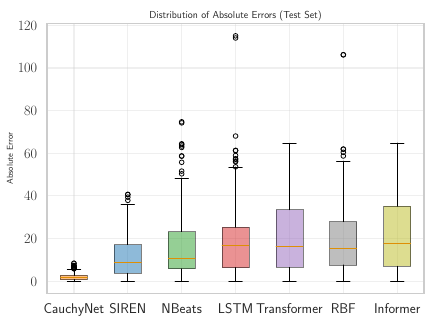}
    \caption{(Left) Training/validation loss for 200 epochs. CauchyNet converges fastest and stays lower
than baselines. (Right) Absolute errors on the test set: CauchyNet has the smallest median and tightest
spread (MAE of 2.3 vs. 10+ for most baselines).}
\end{figure}

\begin{figure}[H]
    \centering
    \includegraphics[scale=1]{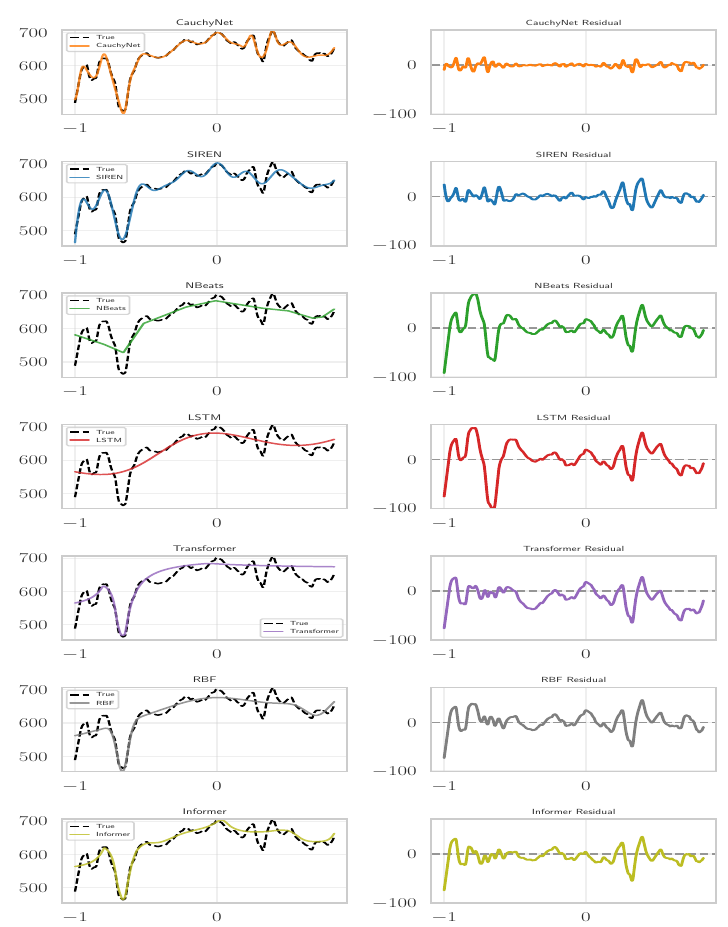}
    \caption{Predicted versus true function and residuals for the 1D imputation task: (Left
column) Predicted curves for each model compared to the ground truth (dashed line). (Right column)
Residuals near the sharp spikes, highlighting CauchyNet's close alignment with the true function compared
to overshooting or flattening observed in baselines like SIREN and N-BEATS.}
\label{fig_predictedversustrue}
\end{figure}

\clearpage
Fig.~\ref{fig_predictedversustrue} presents each model's predicted curve on a dense grid within [-1, 1], focusing on residuals
near the spikes. CauchyNet aligns closely with the ground truth, whereas all baselines either
overshoot or flatten out near singularities.
CauchyNet's holomorphic bias and reciprocal activation yield robust forecasts with minimal data.
Despite increasing data numbers (700), it outperforms large models like N-BEATS and Transformers
in accuracy and parameter efficiency.

\subsection*{Experiment 5: Ablation Study and Parameter Sensitivity}
We investigate how CauchyNet's architectural components (e.g., elliptical initialization, imaginary
penalty) and hyperparameters affect its performance. We use the same setup as in Experiment 1 and the same dataset.

\noindent\textbf{Penalizing the imaginary component.} According to Section 4, we know that the CauchyNet returns two separate tensors corresponding
to the real and imaginary parts of the output o. Let $y_{true}$ denote the target output. The model produces a complex output $o=\Re(o)+i\Im(o)$. According to (5), the loss function is:
\begin{align}
\mathcal{L}\;=\;\left(\Re(o)\,-\,y_\mathrm{true}\right)^{2}\,+\,\lambda\,\left|\Im(o)\right|^{2},\label{eq_loss_func}
\end{align}
where $\lambda > 0$ is a hyperparameter that penalizes the magnitude of the imaginary component, encouraging $\Im(o)\approx0$ during training.

We let the hyperparameter $\lambda$ varies, and takes the values from the set $\{0.1, 0.3, 0.5, 1, 1.5\}$ in the Loss function (\ref{eq_loss_func}).

\begin{figure}[H]
    \centering
    \includegraphics[scale=0.1]{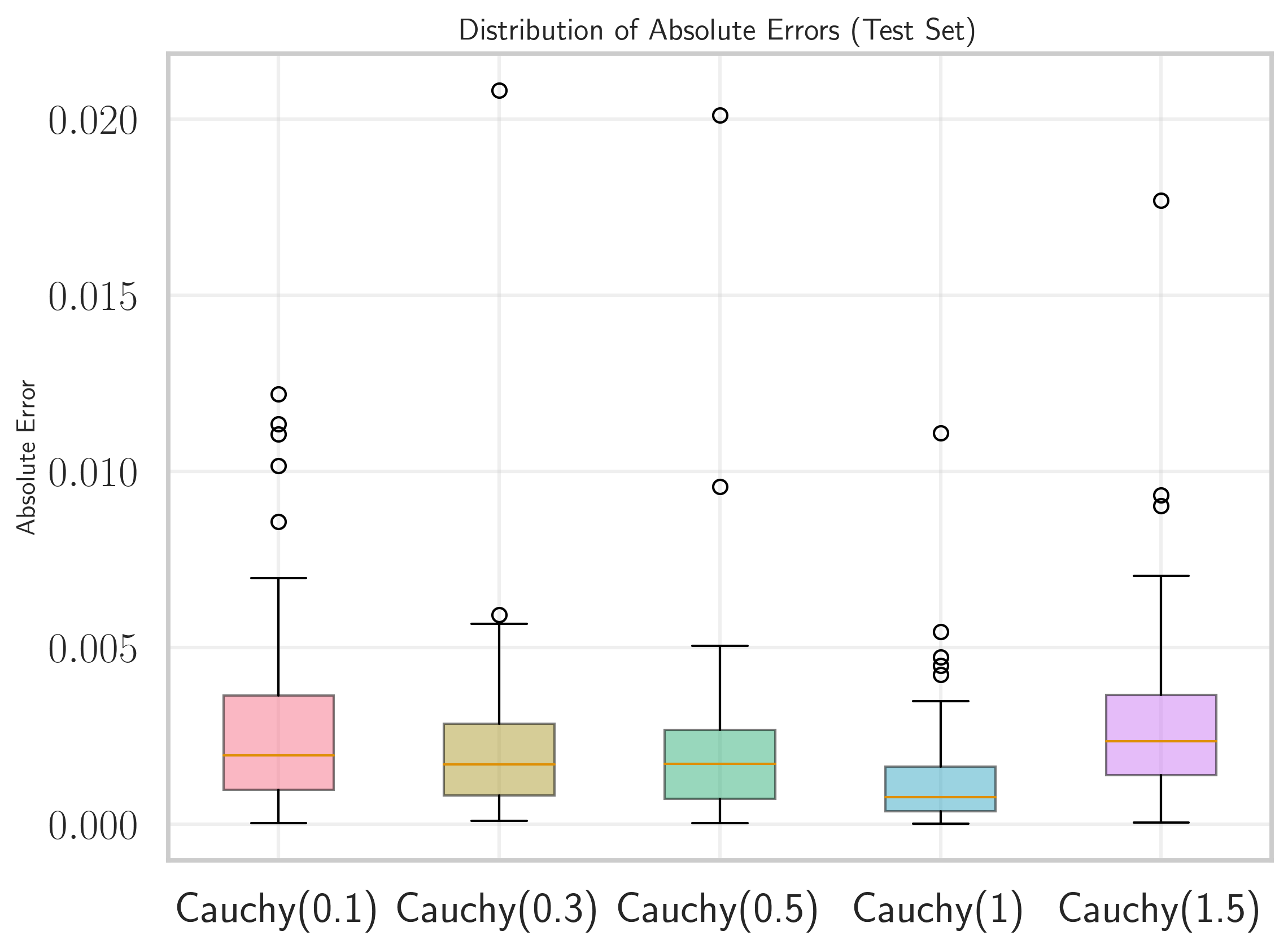}
    \caption{Box plot of the test performance (MSE loss in log scale) across 200 epochs for the
jyperparameter $\lambda$ taken values in ${0.1, 0.3, 0.5, 1, 1.5}$, which defines the CauchyNet variants, denoted
as Cauchy($\lambda$). Retaining the imaginary component in the output and incorporating it into the loss
with $\lambda = 1$ improves test MSE.}
\label{figure_boxplot_testperformance}
\end{figure}

Fig.~\ref{figure_boxplot_testperformance} illustrates the box plot of the test loss for CauchyNet with different values of hyperparameter $\lambda$. We observe that properly adding the imaginary part of the output as a regularization term tends to lower the loss, consistent with the hypothesis that the imaginary dimension can serve as a valuable latent parameter space.

The inclusion of an imaginary penalty is critical for stabilizing training and maintaining low errors.
Removing this penalty leads to performance degradation, highlighting its importance in the model's architecture.

\noindent\textbf{Parameter Sensitivity.} Understanding how different hyperparameters affect CauchyNet's performance is crucial for optimizing
its functionality. To this end, we conducted a series of experiments varying hidden dimensions, data sizes, learning rates, and weight decay values. Heatmaps in Figures~\ref{fig_training_validation_loss_for200} show test MSE over these ranges. Larger hidden sizes consistently lower MSE, and an Adam learning rate of 0.01 with weight decay $1\times10^{-5}$ generally works best.

\begin{figure}
    \centering
    \includegraphics[scale=0.12]{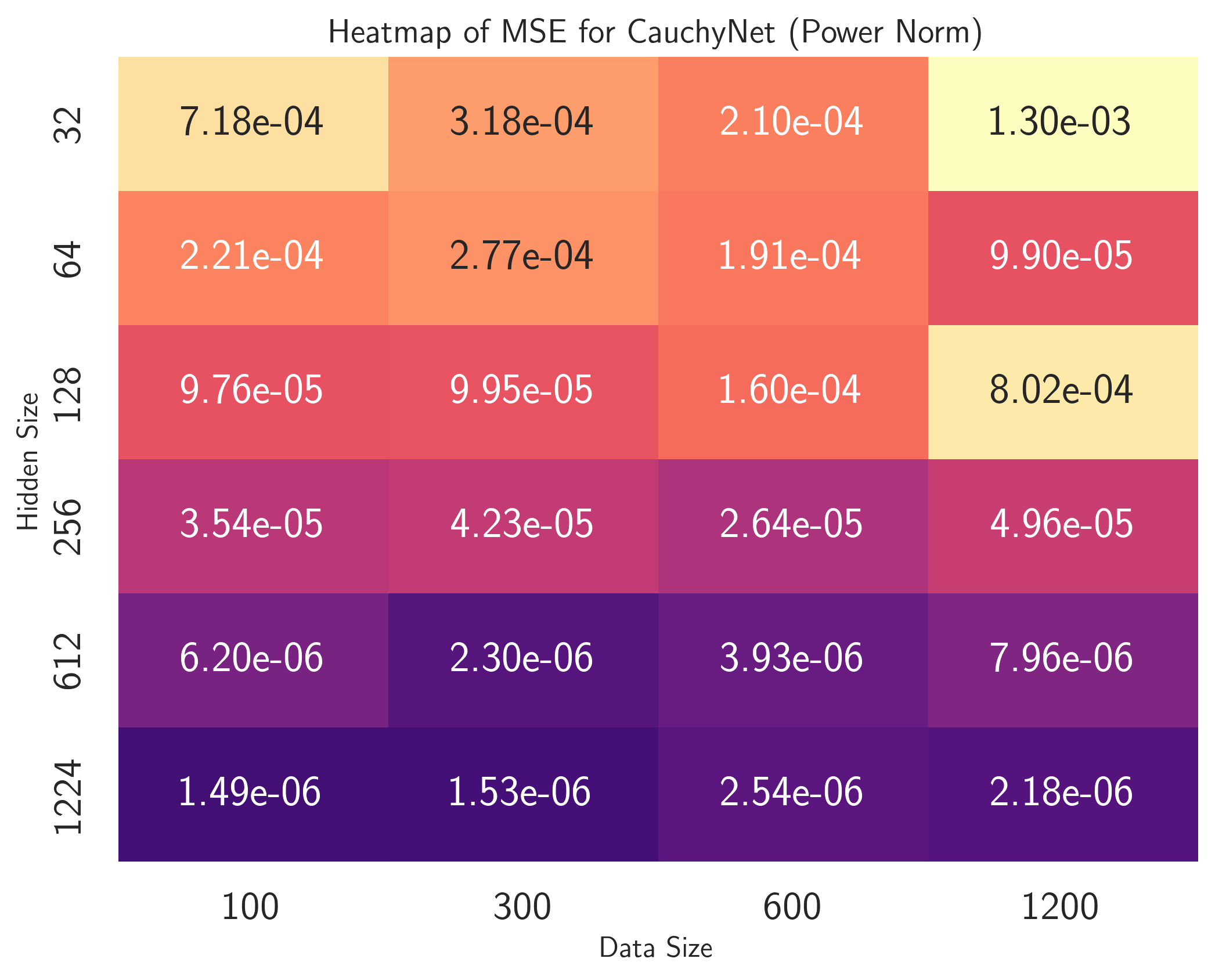}
    \includegraphics[scale=0.12]{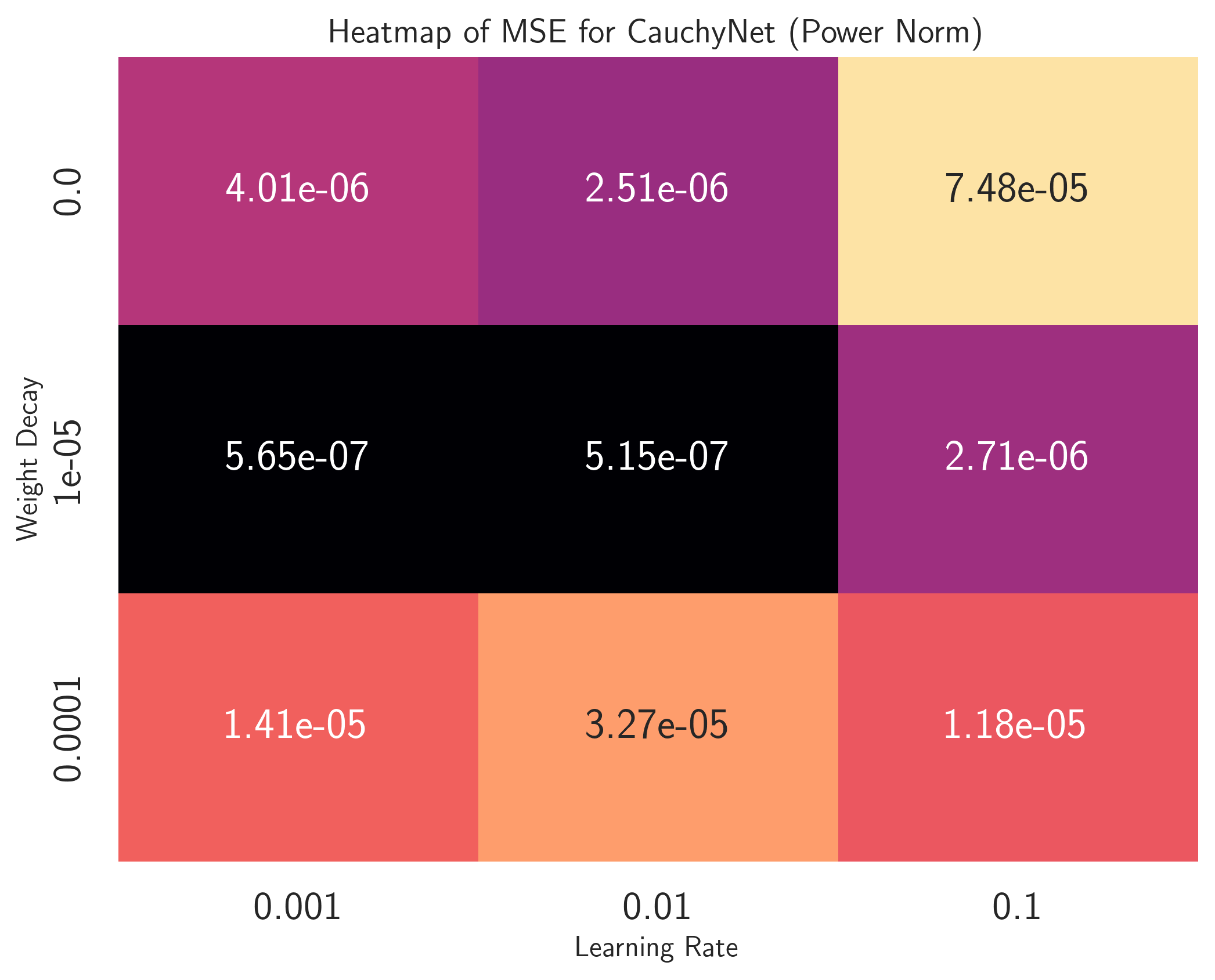}
    \caption{(Left) Heatmap of test MSE for CauchyNet across varying hidden dimensions (32, 64, 128, 256, 612,
1224) and dataset sizes (100, 300, 600, 1200). Larger models can achieve MSE as low as $10^{-6}$, underscoring
strong scalability. (Right) Test MSE under learning rates {0.001, 0.01, 0.1} and weight decay ${0.0, 1 × 10^{-5}, 1 × 10^{-4}}$.
The combination $(0.01, 1 × 10^{-5})$ is most stable.}
\label{fig_training_validation_loss_for200}
\end{figure}

Ablations confirm that imaginary-part regularization and a rational activation are central to CauchyNet's performance. Parameter sensitivity results show that the model scales efficiently with hidden dimension and can be robustly tuned for diverse data sizes or complexity. 

\noindent\textbf{Overall Summary of Experiments.} Across multiple tasks, including 1D function approximation, 2D surface modeling, time-series forecasting, and missing data imputation, CauchyNet consistently outperforms or rivals large baselines in accuracy, speed, and parameter efficiency. Its rational-based, holomorphic design supports stable training under limited data, near singular features, and real-world constraints, making it a promising candidate for low-resource deployments and complex scientific problems.

\section{Proof of Cauchy Approximation Theorem}
The proof leverages the \emph{multivariate Cauchy's integral formula}, a cornerstone in complex analysis, which provides a way to express an analytic function inside a domain exclusively in terms of its boundary values.

Given a domain $U=\prod_{i=1}^N U_i$, suppose $f: M \rightarrow \mathbb{R}$ is extended to an analytic function $\bar{f}$ on $U$, and is continuous on its closure $\bar{U}$. Then, the \emph{multivariate Cauchy's integral formula} is articulated for all $\boldsymbol{z}=\left(z_1, \ldots, z_N\right) \in \dot{U}$ as:

\begin{align}
\bar{f}(\boldsymbol{z})=\frac{1}{(2 \pi i)^N} \int_{\zeta_1 \in \partial U_1} \cdots \int_{\zeta_N \in \partial U_N} \frac{\bar{f}(\zeta)}{\left(\zeta_1-z_1\right) \cdots\left(\zeta_N-z_N\right)} d \zeta_1 \ldots d \zeta_N
\end{align}

where $\zeta=\left(\zeta_1, \ldots, \zeta_N\right)$.

To approximate $\bar{f}(\boldsymbol{z})$, we discretize the boundary $\partial \bar{U}=\prod_{i=1}^N \partial \bar{U}_i$ into a mesh $\mathcal{M}_m$ with mesh size $D_m=\max \left\{\operatorname{diam}\left(e_{\boldsymbol{k}}\right)\right\}$, where $e_{\boldsymbol{k}}$ are the partition elements of the mesh. As $m \rightarrow \infty, D_m \rightarrow 0$.
Define a sequence of meshes $\left\{\mathcal{M}_m\right\}_{m \in \mathbb{N}}$ such that each mesh $\mathcal{M}_m$ partitions $\partial \bar{U}$ into elements $\left\{e_{\boldsymbol{k}}\right\}$, where $\boldsymbol{k}=\left(k_1, \ldots, k_N\right), k_i=1,2, \ldots, n_i$, and $\prod_{j=1}^N n_j=m$. The sigma-algebra $\mathcal{F}_m$ is generated by the partition $\mathcal{M}_m$, forming a filtration $\left\{\mathcal{F}_m\right\}$ with $\mathcal{F}_m \subset \mathcal{F}_{m+1} \subset \mathcal{B}(U)$, where $\mathcal{B}(U)$ is the Borel sigma-algebra on $U$.

Consider $h(\boldsymbol{z})=\bar{f}(\boldsymbol{z})$ defined by \emph{Cauchy's integral formula}. The conditional expectation $\mathbb{E}\left(h \mid \mathcal{F}_m\right)$ approximates $h(\boldsymbol{z})$ by averaging over the mesh elements. Specifically, for each $\boldsymbol{z} \in U$ :

\begin{align}
h_m(\boldsymbol{z})=\mathbb{E}\left(h \mid \mathcal{F}_m\right)=\frac{1}{(2 \pi i)^N} \sum_{\boldsymbol{k}} \frac{\bar{f}\left(\zeta_{\boldsymbol{k}}\right) \cdot m\left(e_{\boldsymbol{k}}\right)}{\left(\zeta_{\boldsymbol{k}, 1}-z_1\right) \cdots\left(\zeta_{\boldsymbol{k}, N}-z_N\right)},\notag
\end{align}

where $\zeta_{\boldsymbol{k}}=\left(\zeta_{\boldsymbol{k}, 1}, \ldots, \zeta_{\boldsymbol{k}, N}\right) \in e_{\boldsymbol{k}}$, and $m\left(e_{\boldsymbol{k}}\right)$ is the Lebesgue measure of the mesh element $e_{\boldsymbol{k}}$.
Setting $\theta_{\boldsymbol{k}}=\frac{\bar{f}\left(\zeta_{\boldsymbol{k}}\right) \cdot m\left(e_{\boldsymbol{k}}\right)}{(2 \pi i)^N}$, we obtain:

\begin{align}
h_m(\boldsymbol{z})=\sum_{\boldsymbol{k}} \frac{\theta_{\boldsymbol{k}}}{\left(\zeta_{\boldsymbol{k}, 1}-z_1\right) \cdots\left(\zeta_{\boldsymbol{k}, N}-z_N\right)}.\notag
\end{align}
This expression represents $h_m(\boldsymbol{z})$ as a linear combination of Cauchy kernels centered at the mesh points $\zeta_{\boldsymbol{k}}$.

As $m \rightarrow \infty, D_m \rightarrow 0$, and the approximation $h_m(\boldsymbol{z})$ converges to $h(\boldsymbol{z})$ uniformly on compact subsets of $U$. This follows from the Dominated Convergence Theorem and the properties of the Cauchy Integral Formula.

Moreover, since $\left\{h_m\right\}$ forms a martingale with respect to the filtration $\left\{\mathcal{F}_m\right\}$, by the Martingale Convergence Theorem, $h_m(\boldsymbol{z}) \rightarrow h(\boldsymbol{z})$ almost surely and in $L^2$. Additionally, since $\left\{h_m\right\}$ are analytic functions and uniformly bounded on compact subsets $M \subset \stackrel{\circ}{U}$, Montel's Theorem ensures that the convergence is uniform on $M$.

Therefore, for any $\epsilon>0$, there exists an $m$ such that:

\begin{align}
\left|h(\boldsymbol{z})-\sum_{\boldsymbol{k}} \frac{\theta_{\boldsymbol{k}}}{\left(\zeta_{\boldsymbol{k}, 1}-z_1\right) \cdots\left(\zeta_{\boldsymbol{k}, N}-z_N\right)}\right|<\epsilon,\notag
\end{align}

for all $\boldsymbol{z} \in M$.
This establishes that $h(\boldsymbol{z})$ can be approximated arbitrarily closely by a finite sum of Cauchy kernels, thereby proving Theorem 1.

\section{Introduction to Holomorphic Functions}
Below, we revise the definition and some properties of holomophic functions~\citep{Ahlfors1966}. Holomorphic functions play a pivotal role in complex analysis and have significant applications in various fields, including machine learning. This appendix provides a concise introduction to holomorphic functions, outlining their definitions, key properties, and relevance to the architecture and theoretical foundations of CauchyNet.

A function $f: U \rightarrow \mathbb{C}$, where $U \subseteq \mathbb{C}$ is an open subset of the complex plane, is called \emph{holomorphic} (or \emph{analytic}) at a point $z_0 \in U$ if it is complex differentiable in some neighborhood around $z_0$. Formally, $f$ is holomorphic at $z_0$ if the following limit exists:

\begin{align}
f^{\prime}\left(z_0\right)=\lim _{z \rightarrow z_0} \frac{f(z)-f\left(z_0\right)}{z-z_0}.\notag
\end{align}

If $f$ is holomorphic at every point in $U$, it is said to be holomorphic on $U$.
Holomorphic functions exhibit several remarkable properties that distinguish them from their real-valued counterparts:

Every holomorphic function is analytic, meaning it can be locally represented by a convergent power series. Specifically, for any $z_0 \in U$, there exists a radius $R>0$ and coefficients $a_n \in \mathbb{C}$ such that

\begin{align}
f(z)=\sum_{n=0}^{\infty} a_n\left(z-z_0\right)^n \quad \text { for all } z \text { with }\left|z-z_0\right|<R.\notag
\end{align}

A necessary condition for $f(z)=u(x, y)+i v(x, y)$ to be holomorphic is that its real and imaginary parts satisfy the Cauchy-Riemann equations:

\begin{align}
\frac{\partial u}{\partial x}=\frac{\partial v}{\partial y}, \quad \frac{\partial u}{\partial y}=-\frac{\partial v}{\partial x}.\notag
\end{align}

These equations ensure that the function behaves well under complex differentiation.

Holomorphic functions are infinitely differentiable within their domain of holomorphy. This smoothness facilitates the application of various analytical techniques.

If $f^{\prime}(z) \neq 0$ at a point $z$, then $f$ preserves angles locally at $z$. This property is known as conformality and is essential in fields like fluid dynamics and electromagnetic theory.

\textbf{Liouville's Theorem}: Any bounded entire (holomorphic on all of $\mathbb{C}$ ) function must be constant. This theorem has profound implications, including a proof of the fundamental theorem of algebra.

\textbf{Maximum Modulus Principle}: If $f$ is non-constant and holomorphic on a bounded domain $D$, then $|f(z)|$ cannot attain its maximum value inside $D$. Instead, the maximum occurs on the boundary of $D$.

In the context of CauchyNet, holomorphic functions provide a mathematically robust framework for designing activation functions and network architectures that can handle complex patterns, including near-singularities and oscillatory behaviors. By leveraging properties of holomorphic functions, such as their smoothness and conformality, CauchyNet achieves:\\
- \textbf{Stable Gradient Flow}: Holomorphic activations facilitate stable and efficient backpropagation by ensuring smooth and well-behaved gradients.\\
- \textbf{Expressive Power}: The ability of holomorphic functions to represent a wide range of behaviors with fewer parameters aligns with the goals of parameter efficiency and compactness in CauchyNet.\\
- \textbf{Universal Approximation}: The Cauchy Integral Formula and related approximation theorems underpin the network's capacity to approximate complex functions, even in data-scarce environments.

By embedding holomorphic principles into its architecture, CauchyNet not only draws from classical mathematical theory but also enhances its capability to perform robustly across various challenging tasks, including function approximation, time-series forecasting, and missing-value imputation.

\section{Wirtinger Calculus and Backpropagation in Complex-Valued Neural Networks}

\subsection{Introduction to Wirtinger Derivatives}

In complex analysis and optimization, the Wirtinger derivatives provide a convenient framework for computing derivatives of functions with respect to complex variables. This approach is particularly useful in the context of complex-valued neural networks, where the standard rules of calculus for real-valued functions do not directly apply.

For a complex variable $z=x+i y$, where $x, y \in \mathbb{R}$ and $i$ is the imaginary unit, any complex-valued function $f: \mathbb{C} \rightarrow \mathbb{C}$ can be viewed as a function of two real variables:
\begin{align}
f(z)=f(x+i y)=u(x, y)+i v(x, y),\notag
\end{align}
where $u, v: \mathbb{R}^2 \rightarrow \mathbb{R}$ are the real and imaginary parts of $f$, respectively.

The Wirtinger derivatives are defined as:
\begin{align}
\begin{aligned}
\frac{\partial f}{\partial z} & =\frac{1}{2}\left(\frac{\partial f}{\partial x}-i \frac{\partial f}{\partial y}\right) \\
\frac{\partial f}{\partial z^*} & =\frac{1}{2}\left(\frac{\partial f}{\partial x}+i \frac{\partial f}{\partial y}\right),\notag
\end{aligned}
\end{align}
where $z^*=x-i y$ is the complex conjugate of $z$.

These derivatives treat $z$ and $z^*$ as independent variables, allowing us to compute gradients in a manner analogous to real-valued functions. This is particularly advantageous when dealing with functions that are not holomorphic (i.e., not complex-differentiable in the traditional sense), as it enables the use of gradient-based optimization methods in complex-valued neural networks.
\subsection{Derivative of the Activation Function}

The Cauchy Activation Function in CauchyNet is defined as $\mathscr{X}(\boldsymbol{z})=\prod_{i=1}^N z_i^{-1}$, where $\boldsymbol{z} \in \mathbb{C}_*^N$ (i.e., $z_i \neq 0$ for all $i$ ).

To implement backpropagation, we need to compute the derivative of $\mathscr{X}(\boldsymbol{z})$ with respect to each component $z_j$. Since $\mathscr{X}(\boldsymbol{z})$ is a holomorphic function on $\mathbb{C}_*^N$, we can use standard complex differentiation techniques.

\noindent\textbf{Theorem 6.}\ \textit{The derivative of $\mathscr{X}(\boldsymbol{z})$ with respect to $z_j$ is:}

\begin{align}
\frac{\partial \mathscr{X}(\boldsymbol{z})}{\partial z_j}=-\mathscr{X}(\boldsymbol{z}) \cdot z_j^{-1}.\notag
\end{align}

\textit{Proof.} Consider $\mathscr{X}(\boldsymbol{z})=\prod_{i=1}^N z_i^{-1}=P^{-1}$, where $P=\prod_{i=1}^N z_i$.
The derivative with respect to $z_j$ is:
\begin{align}
\frac{\partial \mathscr{X}(\boldsymbol{z})}{\partial z_j}=\frac{\partial}{\partial z_j}\left(P^{-1}\right)=-P^{-2} \cdot \frac{\partial P}{\partial z_j}.\notag
\end{align}

Compute $\frac{\partial P}{\partial z_j}$, we get $\frac{\partial P}{\partial z_j}=\prod_{\substack{i=1 \\ i \neq j}}^N z_i$.

Therefore,
\begin{align}
\begin{aligned}
\frac{\partial \mathscr{X}(\boldsymbol{z})}{\partial z_j} & =-P^{-2} \cdot \prod_{\substack{i=1 \\
i \neq j}}^N z_i=-\left(\prod_{i=1}^N z_i\right)^{-2} \cdot\left(\prod_{\substack{i=1 \\
i \neq j}}^N z_i\right) \\
& =-\frac{1}{\prod_{i=1}^N z_i} \cdot \frac{1}{z_j}=-\mathscr{X}(\boldsymbol{z}) \cdot z_j^{-1}.\notag
\end{aligned}
\end{align}

Thus, the derivative simplifies to:
\begin{align}
\frac{\partial \mathscr{X}(\boldsymbol{z})}{\partial z_j}=-\mathscr{X}(\boldsymbol{z}) \cdot z_j^{-1}.\notag
\end{align}

Since $\mathscr{X}(\boldsymbol{z})$ is holomorphic, its Wirtinger derivative with respect to $z_j^*$ is zero:

\begin{align}
\frac{\partial \mathscr{X}(\boldsymbol{z})}{\partial z_j^*}=0.\notag
\end{align}

The Wirtinger derivative with respect to $z_j$ is the same as the standard derivative.$\hfill\blacksquare$

\subsection{Gradient Computation in Backpropagation}

In backpropagation, the gradient of the loss function $L$, which includes contributions from both the real part $y$ and the imaginary error term $e$, is computed using Wirtinger derivatives. The regularization term involving $e$ ensures that the imaginary component remains small, aiding the network in stabilizing the gradient magnitudes and improving convergence during training.

When training the network, we need to compute the gradient of the loss function $L$ with respect to the complex parameters $\theta$ (elements of $\mathbf{B}$ and $\mathbf{c}$ ). Using the chain rule and Wirtinger derivatives, the gradient with respect to $z_j$ can be expressed as:
\begin{align}
\frac{\partial L}{\partial z_j}=\frac{\partial L}{\partial \mathscr{X}} \cdot \frac{\partial \mathscr{X}(\boldsymbol{z})}{\partial z_j}.\notag
\end{align}

Since $L$ is real-valued, and $\mathscr{X}(\boldsymbol{z})$ is holomorphic, the Wirtinger derivative simplifies the computation.

\noindent\textbf{Backpropagation through the Activation Function.} Suppose we have a real-valued loss function $L$ dependent on the network output $\mathbf{y}$ :
\begin{align}
L=\mathcal{L}\left(\mathbf{y}, \mathbf{y}_{\text {true }}\right)+\lambda|\mathbf{e}|^2,\notag
\end{align}
where $\mathbf{y}=\Re(o), \mathbf{e}=\Im(o)$, and $o=\mathbf{c}^{\top} \boldsymbol{h}$.
The derivative of $L$ with respect to $o$ is:
\begin{align}
\frac{\partial L}{\partial o}=\frac{\partial L}{\partial y}+i \frac{\partial L}{\partial e}.\notag
\end{align}

Next, the derivative with respect to $\mathbf{c}$ is:
\begin{align}
\frac{\partial L}{\partial \mathbf{c}}=\boldsymbol{h} \cdot \frac{\partial L}{\partial o}.\notag
\end{align}

For each hidden unit $k$, the derivative with respect to $h_k$ is:
\begin{align}
\frac{\partial L}{\partial h_k}=\mathbf{c}_k \cdot \frac{\partial L}{\partial o}.\notag
\end{align}

Using the derivative of the activation function:
\begin{align}
\frac{\partial h_k}{\partial \mathbf{H}_{k, j}}=-h_k \cdot\left(\mathbf{H}_{k, j}+\varepsilon\right)^{-1}.\notag
\end{align}

Therefore, the gradient with respect to the bias parameters $\mathbf{B}_{k, j}$ is:
\begin{align}
\frac{\partial L}{\partial \mathbf{B}_{k, j}}=\frac{\partial L}{\partial h_k} \cdot \frac{\partial h_k}{\partial \mathbf{H}_{k, j}}=-h_k \cdot \mathbf{c}_k \cdot \frac{\partial L}{\partial o} \cdot\left(\mathbf{H}_{k, j}+\varepsilon\right)^{-1}.\notag
\end{align}

Using these gradients, we update the parameters using an optimization algorithm like stochastic gradient descent (SGD) or Adam:
\begin{align}
\theta \leftarrow \theta-\eta \frac{\partial L}{\partial \theta},\notag
\end{align}

where $\eta$ is the learning rate.
\subsection{Advantages of Wirtinger derivatives in CauchyNet}

The use of Wirtinger derivatives in CauchyNet offers several advantages:\\
- \textbf{Computational Efficiency:} By treating $z$ and $z^*$ as independent variables, Wirtinger calculus simplifies the gradient computations, avoiding the need to separate real and imaginary components explicitly.\\
- \textbf{Compatibility with Gradient-Based Methods:} The framework aligns well with standard optimization algorithms like stochastic gradient descent, enabling straightforward integration into existing training pipelines.\\
- \textbf{Handling Non-Holomorphic Functions:} Even though the activation functions in CauchyNet are holomorphic, the overall network may not be due to the inclusion of realvalued loss functions. Wirtinger calculus accommodates such cases effectively.\\
- \textbf{Stable Gradient Flow:} The inversion-based activation function, combined with the imaginary error penalty, ensures that gradients remain stable and do not vanish or explode during training.

When implementing Wirtinger calculus in practice, the following considerations are important: 1) Complex arithmetic can introduce numerical errors, especially when dealing with very small or large magnitudes. Careful handling of numerical precision is essential. Adding a small constant $\varepsilon>0$ to denominators prevents division by zero and mitigates numerical instability. 2) Modern deep learning frameworks like TensorFlow and PyTorch offer support for complex numbers and automatic differentiation using Wirtinger derivatives, facilitating the implementation of complexvalued neural networks. 3) Proper initialization of complex weights and biases is crucial. We adopt a complex Xavier initialization, where both the real and imaginary parts are initialized from a normal distribution with zero mean and variance $\frac{2}{N+h}$, where $N$ is the input dimension and $h$ is the hidden dimension.

Wirtinger calculus provides a powerful tool for optimizing complex-valued functions, making it well-suited for training models like CauchyNet. By enabling efficient computation of gradients, it allows the network to leverage the benefits of complex analysis, such as holomorphic properties and enhanced expressiveness, while maintaining compatibility with gradient-based optimization methods. This facilitates stable and efficient training of complex-valued neural networks, contributing to their effectiveness in tasks involving near-singularities and data-scarce scenarios.

\bibliographystyle{abbrvnat}
\bibliography{ref}

\end{document}